\documentclass[aps,pre,twocolumn]{revtex4-1}
    
    \usepackage{amsmath}
    \usepackage{multirow}
    \usepackage{array}
    \usepackage{graphicx}
    \usepackage{diagbox}
    \usepackage{subfigure}
    \usepackage{caption}
    \usepackage{bbding}
    \usepackage{booktabs} 
    \usepackage{makecell}
    \usepackage[usenames,dvipsnames]{color}
    \usepackage{colortbl}
    \definecolor{mygray}{gray}{0.5}
    \makeatletter
    \newcommand{\thickhline}{%
    \noalign {\ifnum 0=`}\fi \hrule height 1pt
    \futurelet \reserved@a \@xhline
    }
    \newcolumntype{"}{@{\hskip\tabcolsep\vrule width 1pt\hskip\tabcolsep}}
    \makeatother
    
\begin{document}
    \captionsetup[figure]{labelfont={bf},name={Figure},labelsep=period}
    \captionsetup[table]{labelfont={bf},name={Table},labelsep=period}
    \title{Simultaneous Implementation Features Extraction and Recognition Using C3D Network for WiFi-based Human Activity Recognition}
    \author{Yafeng Liu}
    \email{yafeng@cumt.edu.cn}
    \affiliation{Internet of Things Research Center, China University of Mining and Technology, Xuzhou, China}    
    \author{Tian Chen}
    \affiliation{Internet of Things Research Center, China University of Mining and Technology, Xuzhou, China}
    \author{Zhongyu Liu}
    \affiliation{Internet of Things Research Center, China University of Mining and Technology, Xuzhou, China}
    \author{Lei Zhang}
    \email{lei@cumt.edu.cn}
    \affiliation{Xuzhou University of Technology}
    \author{Yanjun Hu}
    \affiliation{School of Information and Control Engineering, China University of Mining and Technology, Xuzhou, China}
    \author{Enjie Ding}
    \affiliation{Internet of Things Research Center, China University of Mining and Technology, Xuzhou, China}
    \date{{\small \today}}
    
    \begin{abstract}
    Human actions recognition has attracted more and more people's attention. Many technology have been developed to express human action's features, such as image, skeleton-based, and channel state information(CSI). Among them, on account of CSI's easy to be equipped and undemanding for light, and it has gained more and more attention in some special scene. However, the relationship between CSI signal and human actions is very complex, and some preliminary work must be done to make CSI features easy to understand for computer. Nowadays, many work departed CSI-based features' action dealing into two parts. One part is for features extraction and dimension reduce, and the other part is for time series problems. Some of them even omitted one of the two part work. Therefore, the accuracies of current recognition systems are far from satisfactory. In this paper, we propose a new deep learning based approach, i.e. C3D network and C3D network with attention mechanism, for human actions recognition using CSI signals. This kind of network can make feature extraction from spatial convolution and temporal convolution simultaneously, and through this network the two part of CSI-based human actions recognition mentioned above can be realized at the same time. The entire algorithm structure is simplified. The experimental results show that our proposed C3D network is able to achieve the best recognition performance for all activities when compared with some benchmark approaches.
    \end{abstract}

    \maketitle
    
    \section{Introduction}
    Human action recognition concerns the task of automatically interpreting a sample sequence to decide what action or activity is being performed by the subjects in the scene \citep{journals/csur/AggarwalR11, journals/IEEE/Ji2013, journals/NN/Schmidhuber201585, journals/FRAI/Vrigkas2015, journals/IVC/Herath20174}. It is a relevant topic in artificial intelligence, with practical applications such as video surveillance, human-computer interaction, gaming, sports arbitration, sports training, smart homes, life-care systems, among many others \citep{journals/PRL/Wang2013, journals/AIR/Rautaray2015, DBLP:journals/tist/WangWZGNJZL16}. As we know, there are various behavior recognition modalities based on different types of signals. Camera and video are the most traditional and important approach for recording human activities \citep{journals/IEEE/Yang2017, journals/IEEE/Mario2019}. Moreover, optical flow and skeleton sequence are always also employed as kinds of action recording tools \citep{journals/IEEE/Hsieh2018, journals/IEEE/Yang2019}. However, all of methods mentioned above have the fundamental limitations of requiring enough lighting and demanding personal privacy protecting. Wearable sensors are another popular for human activity recognition due to the high recognition accuracy \citep{journals/AFM/amjadi2015}. But the systems based on wearable sensors require users to take extra devices for activity recognition, which is inconvenient and obstructive for users. Otherwise, some sensors, such as accelerator, gyroscope and barometer, can be embedded in mobile terminal, and the terminal can be treated as a sensing platform for human activity recognition \citep{journals/IEEE/chen2017}. Some weakness also exists here. So many sensors running in terminal will increase their battery usage, which means in some cases they are restricted by the amount of battery power.
    
    Comparing with conventional human activity recognition signal, Wireless sensing technology is a new sensing mechanism that does not require sensor devices to be installed or attached to the target object \citep{journals/IEEE/WangLSLL17}. It is also known as a device-free sensing technology. It makes human activity signal collecting more convenient. Intuitively, when a person resides in the surrounding area of a WiFi transceiver pair, his/her body movement will affect the travel-through WiFi signals. Through analysing such signal characteristics as coarse-grained received signal strength indicator(RSSI) and fine-grained channel state information(CSI), different activities can be recognized. RSSI can be simply accepted by commercial receiver device, such as mobile phone. however RSSI accepted by current signal receiving devices is not very stable, and is easily affected by the environment leading to spurious detections. For this reason, it's not suitable for human activity recognition \citep{journals/IEEE/Gu2016}. The most recent device-free human motion and activity recognition studies are based on CSI instead, because CSI outperforms RSSI in its diversity and stability. CSI is a kind of WiFi owning a more informative characteristic, which has attracted more and more attention due to the abundant and stable information in it \citep{proceeding/ACM/wang2015, DBLP:journals/access/WangJHHDZG19, journals/Sensors/Al-qaness2019, Journals/ACM/Ma2019}.
    
     CSI is more fine-grained information being tracked by Multiple-input multiple-output(MIMO) communications \citep{journals/IEEE/zhang2019, inproceedings/IEEE/Abdelnasser2015, journals/IEEE/Wang2017, journals/Mobicom/Wang2014, DBLP:conf/mobisys/VirmaniS17}. It contains amplitude and phase measurements separately for each orthogonal frequency-division multiplexing(OFDM) subcarriers. To collection CSI information, we need to set up a transceiver device. This device consists of the most common homes/offices WiFi devices(with one antenna) and any PC/laptop(with three antennas). The homes/offices WiFi devices act as the transmitter, and the PC/laptop act as the receiver. A person act in the certain area of a WiFi transceiver pair, his/her body movement will affect the travelthrough Wifi signals. Through analysing CSI signal characteristics, different activities can be recognized. However, the raw CSI signals corresponding to human activity are hard to obtain, since the signals are always affected severely by the wall and indoor physical environment, such as reflection, diffraction, and scattering. To overcome this problem, we use a convolution network to extract the key feature. We arrange several sequential CSI signals as a data matrix, we call it 'a CSI image' here, and an integrated action is made up by such images. So we change this problem into a video classification problem. An 3D convolution network will be employed to solve this problem.
     
    Deep learning has developed amount of network dealing with spatio-temporal event, such as long-short term memory(LSTM) \citep{journals/Sensors/Ordonez2016, journals/IEEE/Ullah2019}, two-stream method \citep{journals/IEEE_conf/wang2017} and 3D convolution network(3D-CNN) \citep{journals/Pattern/YANG20191, journals/Pattern/NUNEZ201880}. Different from convolution network, LSTM does not possess feature extraction ability. Some other methods, such as principal component analysis(PCA) and discrete wavelet transform(DWT), are needed to extract useful CSI information from CSI signal. For a CSI image, it's hard to give out a clear definite to its optic flow, and two-stream methods should be done some changes before it suit for these samples. We don't discuss this here. Therefore, we employ 3D-CNN network for dealing with our problems. 3D-CNN use a form of spatio-temporal convolution means extracting sample features from images sequential. The convolution kernel is a three-dimensional data square. Two dimension of the square is spatial convolution, and the remaining dimension is temporal convolution for time series. The spatial convolution is used for training network for denoising of CSI signals. Images contains redundant information. For example, if we want to make faces recognition, we need extract faces information from an image. The background, light ray and human body is useless, and these information can be thrown away by well trained convolution network. Here, we regard signals affected by the wall and indoor physical environment as useless information. With the help of nonlinear transform of convolution network, the useful raw CSI information can be extracted from CSI images. The spatial attention mechanism is also employed here to get higher accurate results. 
    
    The main contributions of this study can be summarized as follows:
    
    1. To the best of our knowledge, our approach is one of the first to integrate 3D convolution network for exploring CSI temporal signals for action recognition. Different from LSTM approaches, we can use 3D-CNN network to extract useful raw CSI information from original signal without preprocessing.
    
    2. Soft attention mechanism in the spatial convolution and temporal convolution make our recognition processing focus on important CSI signals fragments. Integrated with 3D-CNN, the entire system benefits from not only the comprehensive entire CSI signals, but also local CSI signals that are relevant to actions.

    3. Different from the input in LSTM, we carve the sequential CSI signals of action samples up, and arrange them into several windows. In the way, the CSI arrays are changed into sorts of video-like data, and then they can be dealt by 3D-CNN network.
    
    The rest of this paper is organized as follows. Section II lists out some related work of this motivation. We describe the preprocessing scheme and the 3D-CNN based recognition approach in section III. The implementation and evaluation are presented in section IV. We make the discussion and conclusion in the last section.
    \section{Related Work}
    We present behavior recognition based on WiFi signal in this part and introduce the contents from signal types and their corresponding applications. Also, we introduce how deep learning algorithm developed to dealing with this kind of signal. 
    \subsubsection{channel state information}
    Unlike other WiFi devices, CSI cannot be directly measured with commercial off-the-shelf(COTS) devices. They must be measured by modifying the device driver on Intel 5300 network interface card (NIC), Atheros 9580, and Atheros 9390. With these drivers' help, CSI signals are widely available, many CSI based human activity recognition systems have been developed in the literature, such as Wi-Vi\citep{inproceedings/IEEE/hu2017}, E-eyes\citep{journals/Mobicom/Wang2014}, Wihear\citep{jounals/IEEE/wang2016}, CARM\citep{inproceedings/Mobicom/wang2015}, Wigest\citep{inproceedings/IEEE/Abdelnasser2015}, RT-Fall\citep{journals/IEEE/Wang2017} and others. wang et al.\citep{jounals/IEEE/wang2016} used specialized directional antennas to obtain CSI variations caused by lip movement for recognizing spoken words. They detected and analysed CSI signals reflections from moth movements, and achieved high accuracy for people talking without line-of-sight. Wang et al.\citep{journals/IEEE/Wang2017} presented a real-time, contactless, low-cost indoor fall detection using the commodity WiFi devices called RT-Fall. The system can recognize the fall in the condition that numerous daily activities are performed. Wu et al.\citep{journals/IEEE/wu2019} proposed a TW-see system for human activity recognition. The system can realize the detection of human action through the wall without any dedicated device. 
    
    The problem of activity recognition is somewhat similar to the speech recognition process, and it's also time series sample, essentially. Recently, people have developed many deep learning network fitting for sequential data, such as recurrent neural network(RNN), long-short term memory(LSTM), 3D convolution network(3D-CNN)\citep{journals/IEEE/Greff2017, journals/IEEE/Ji2013}. Some deep learning network are also employed in dealing with CSI temporal signals. There are two main means used before: one is arranging CSI signal into an image-like matrix, and extracting its feature from convolution neural network(CNN) \citep{journals/IEEE_trans/wang2017, journals/IEEE/Gao2017}. The other one is direction making CSI temporal signals as LSTM's inputting, and obtaining the action classification from LSTM's output. Wang et al.\citep{journals/IEEE_trans/wang2017} proposed a sparse autoencoder network, and CSI signal features can be learned automatically from the network. Another network is needed to estimate location, activity, and gesture recognition through these features. Gao et al.\citep{journals/IEEE/Gao2017} transformed CSI temporal signals from different channels into images, and learned these images features with the help of deep learning network. In the end, a softmax regression was employed for localization and activity classification. Chen et al.\citep{journals/IEEE/chen2019} tried Bidireciton LSTM(BLSTM) on the raw CSI temporal signals, and attention mechanism is simultaneously added into the network for higher accurate.
    
    Both of the CNN and LSTM networks have their own advantage for dealing CSI temporal signals, nevertheless some limitations are exist in these two methods. For CNN, the temporal features are integrated into spatial convolution process. The temporal features are important for action recognition. Mixing spatial and temporal features in one convolution kernel make training step confuse, and the result may be affected. For LSTM, lacks of features extraction will be needed another work denoising and dimension reduction. For this reason, some other network needed to be employed for increasing accuracy. Based on these considerations, we will employ 3D-CNN for CSI temporal signal spatial and temporal features extraction. Different from 2D convolution kernel, 3D convolution kernel is formed by two dimensional spatial dimension and one dimensional temporal dimension. Therefore, 3D-CNN can extract spatial features and temporal features separately simultaneously. In this way, we can realize end-to-end action recognition based on CSI temporal signal. Some related work about 3D-CNN is introduced follow.

    \subsubsection{action recognition}
    Nowadays, There are four main kind of action recognition methods based on sequential sample. They are respectively traditional methods, CNN based methods, RNN based methods, and 3D-CNN based methods. Traditional methods usually utilize image's pixel relationship, and extract image features from them. It commonly includes spatial-temporal interest points(STIPs), histogram of gradients(HOG), histogram of optical flows(HOF) and so on\citep{book/princeCVMLI2012}. Inspired by the great success of deep learning models in computer vision tasks, many deep learning networks have been proposed for action recognition. The left three methods are exactly based on three different deep learning networks. 
    
    CNN models\citep{journals/IEEE_conf/wang2017, inproceedings/IEEE/Karpathy2014} are very good at extracting spatial appearance features from subjects and backgrounds, but they are difficult for extracting temporal features. For this reason, CNN based methods should learn motion features utilizing another means, and then fuse with spatial features as action expression. The most frequently used CNN based methods are slow fusion model\citep{inproceedings/IEEE/Karpathy2014} and two-stream model\citep{journals/IEEE_conf/wang2017}. Slow fusion model extends the connectivity of all convolutional layers in time and computes activations through temporal convolutions in addition to spatial convolutions; while two-stream model learns spatial features and motion features from two individual networks, namely SpatialNet, which is trained on single RGB frame and TemporalNet, which is trained on consecutive flow frames. The confidence scores of SpatialNet and TemporalNet are fused to classify actions. Some other improved models based on Two-stream model are also always used as action recognition methods, such as temporal segment network(TSN), trajectory-pooled deep convolutional descriptors(TDD) and so on.
    
    RNN models\citep{journals/IEEE/yousefi2017} can not only record current observation but also store past information by hidden state, through which they can be effective in capturing temporal information. Therefore, RNN models are widely applied in action recognition to capture the motion features in videos. The general process for GNN based methods is that CNN-like models are needed to extract frame-wise features before videos are input into RNN models. The common used RNN based methods is LSTM and gated recurrent unit(GRU). According to the general process, RNN based methods can extract spatial-temporal features utilizing frame-wise features. A soft attention mechanism is sometimes used for key features extraction throughout the video sequence.
    
    3D-CNN model\citep{journals/Pattern/YANG20191, DBLP:journals/corr/TranBFTP14, DBLP:journals/corr/CarreiraZ17, DBLP:journals/corr/abs-1711-11248} extends a 2D convolution to the temporal domain, to extract spatial-temporal featrues for action recognition. It gives consideration to spatial features extraction and temporal features extraction. By this means, one side we can deal with csi signals denosing and dimensional reduction, and the other side we can deal with sequential signals' temporal information. The CSI sequential signals are taken as input directly and we don't rely on any preprocessing. Considering additional kernel dimension having more parameters than 2D CNN and RNN models, some pre-training models will be borrowed for our networks.
    
    \section{Methods}
    The main technique used in this article are WiFi-based sensing technology and 3D convolution neural networks. CSI is considered the new trending metric in WiFi-based sensing technology. With the help of commodity wireless Network Interface Controllers(NICs), CSI signals can be extracted. CSI is a kind of collecting information that describes how wireless signals propagate from the transmitter to the receiver.
    
    \subsection{Channel State Information}
    \begin{figure}[htbp]
    \centering
    \includegraphics[width=0.4\textwidth]{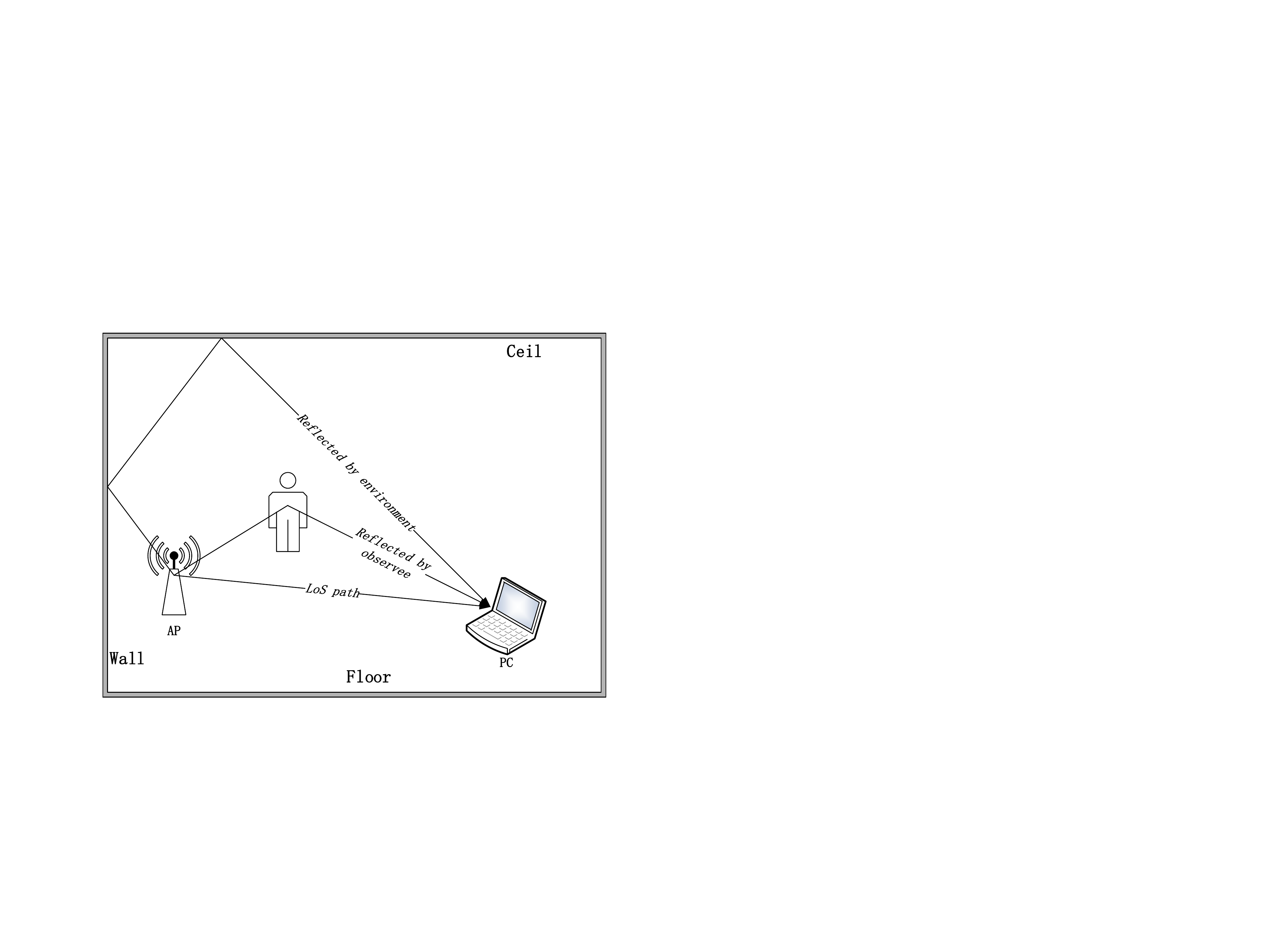}
    \caption{CSI Action signals acquisition device}
    \label{csi_model}
    \end{figure}
    In the IEEE 802.11n/ac standards, CSI is measured and parsed with MIMO and OFDM technology\citep{journals/IEEE/WangLSLL17}. We suppose that there are $n_t$ transmitted antenna, and based on the MIMO model the received signal of the j antenna $y_j(t)$ can be expressed in mathematics form:
    \begin{equation}
    \label{MIMO}
    \begin{split}
    y_j(f,t) &= \sum_{i=1}^{N_{T_x}}h_{i,j}(f, t)*x_i(t) + \eta_j(f, t), \\
         i &= 1,2,\cdots,n_t;j=1,2,\cdots,n_r
    \end{split}                      
    \end{equation}
    where $H(i,j)(f,t)$ is the complex valued channel frequency response(CFR) for carrier frequency f measured at time t between the transmitted antenna i and received antenna j, which is representing CSI information. $X(i)$ is the transmitted signals of the antenna i, and $eta_j(t)$ is an additive white Gaussian noise. Next work is totally based on these CFR values. Let $N_{T_x}$ and $N_{R_x}$ represent the number of transmitting and receiving antennas, respectively. Each CSI signals contains 30 selected OFDM subcarriers between an antenna pair, therefore each CSI signals contains 30 matrices with dimensions $N_{T_x}\times N_{R_x}$. Each entry in these matrix is a CFR value at a certain OFDM subcarrier frequency at a particular time. In conclusion, the dimension of a time-series of entire CSI signal is $30\times N_{T_x}\times N_{R_x}$. 
    \begin{figure*}[htbp]
    \centering
    \includegraphics[width=1.0\textwidth]{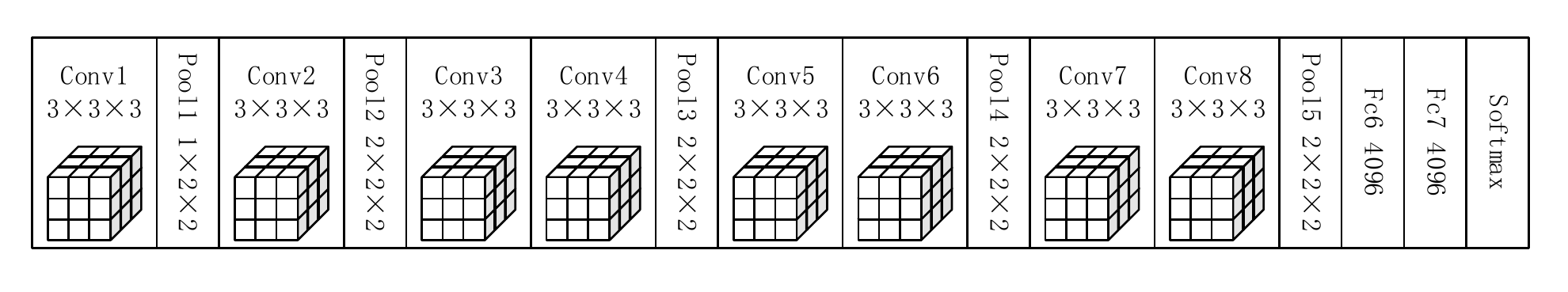}
    \caption{Phasor represent: The dynamic component shows the subcarrier signals between sender and receiver affected by persons with different moving speed(in the magnitude and direction). The total CFR power is the sum of static component and dynamic component.}
    \label{C3D_model}
    \end{figure*}    
    
    \begin{figure}[htbp]
    \centering
    \includegraphics[width=0.4\textwidth]{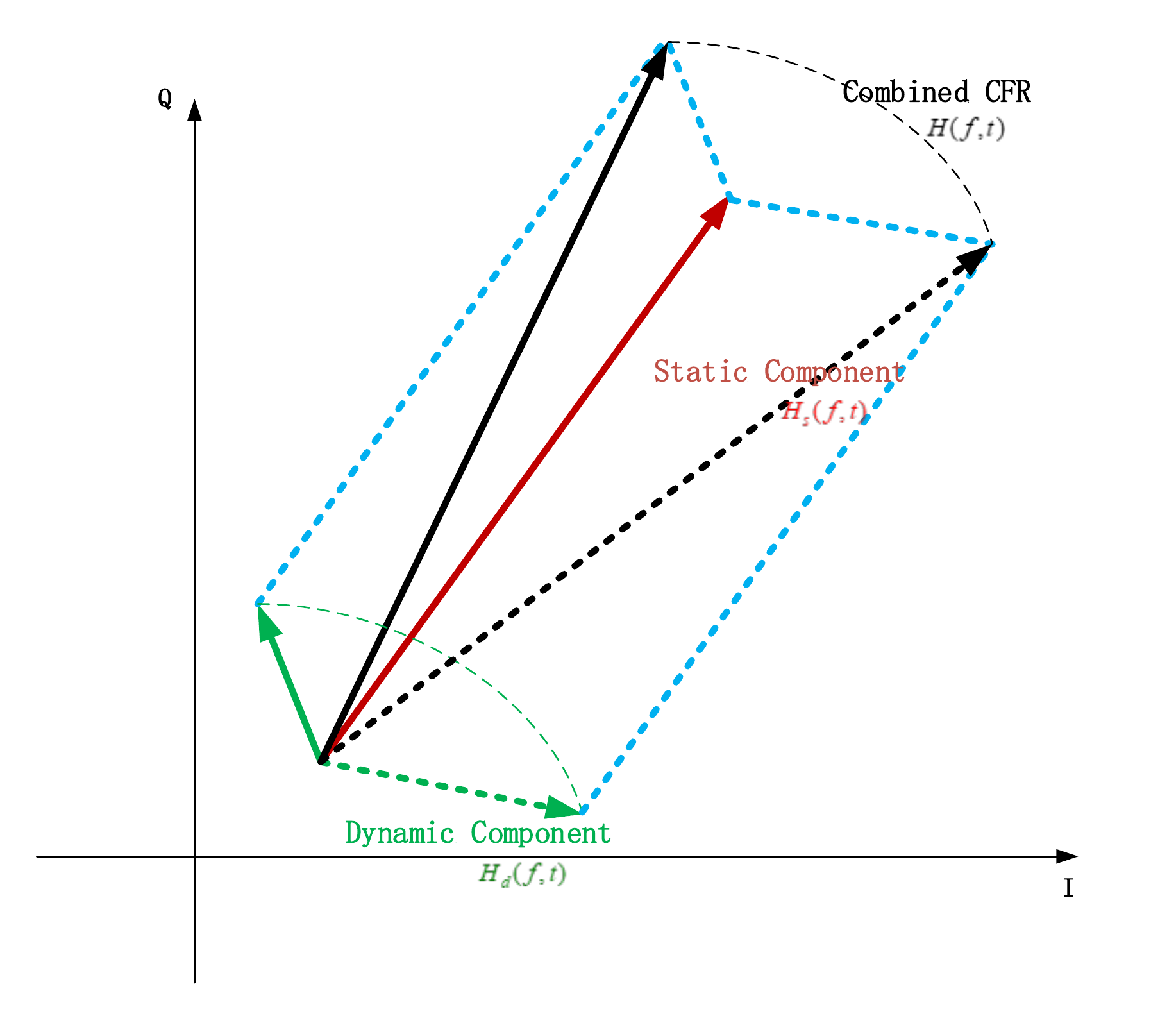}
    \caption{Phasor represent: The dynamic component shows the subcarrier signals between sender and receiver affected by persons with different moving speed(in the magnitude and direction). The total CFR power is the sum of static component and dynamic component.}
    \label{CFR_power}
    \end{figure} 
    The wireless transmitted signals arrives at the receiver through multiple paths, including the LOS path and paths reflected by surrounding objects. Supposed there are N different paths, $H(f,t)$ can be given as follows without the additive noise:
    \begin{equation}
    \label{CFR1}
    H(f,t)=\exp^{-j2\pi\Delta ft}\sum_{k=1}^N a_k(f,t)\exp^{-j2\pi f\tau_k(t)}
    \end{equation}
    where $a_k(f,t)$ and $\tau_k(t)$ are the complex channel attenuation and the time of flight for the $k^{th}$ path, respectively. $\exp^{-j2\pi \Delta ft}$ is phase shift caused by the carrier frequency difference between sender and the receiver.
       
    When a person moves in the CSI signals propagation region, the frequency observed at the receiver will be changed. This is called the Doppler effect. According to the effect, We divide the CFR into two categories such as static CFR and dynamic CFR as in figure \ref{CFR_power}. Dynamic CFR, represented by $H_d(f,t)$, is the sum of paths affected by human moving, and is given by $H_d(f,t)=\sum_{k\in P_d}a_k(f,t)\exp^{-j2\pi d_k(t)/\lambda}$, where $P_d$ is the set of paths affected by human moving. Using $H_s(f)$ for static CFR, Eq.\ref{CFR1} can be modified as
    \begin{equation}
    \label{CFR2}
    H(f,t) = \exp^{-j2\pi\Delta ft}(H_s(f) + \sum_{k\in P_d}a_k(f,t)\exp^{-j\frac{2\pi d_k(t)}{\lambda}})
    \end{equation}
    We suppose that every movement in the signal propagation region is very slow, and therefore can be kept at a constant speed $v_k(t)$. The length of the $k^{th}$ path at time t can be written as $d_k(t) = d_k(0) + v_k(t)$. Then the instantaneous CFR in time t can be defined as:  
    
    \begin{widetext}
    \begin{equation}
    \label{CFR_Power}
    \begin{split}
    &|H(f,t)|^2=\sum_{k\in P_d} 2|H_s(f)a_k(f,t)|cos(\frac{2\pi v_kt}{\lambda}+\frac{\pi d_k(0)}{\lambda}+\phi_{sk}) + \sum_{\substack{k,l\in P_d \\ k\neq l}} 2|a_k(f,t)a_l(f,t) \\
    &\times cos(\frac{2\pi(v_k-v_l)t}{\lambda}+\frac{2\pi(d_k(0)-d_l(0))}{\lambda}+\phi_{kl})+ \sum_{k\in P_d} |a_k(f,t)|^2 + |H_s(f)|^2
    \end{split}
    \end{equation}
    \end{widetext}
    
    where $2\pi(d_k(0)-d_l(0))/\lambda+\phi_{kl}$ and $2\pi d_k(0)/\lambda+\phi_{sk}$ are constants representing initial phase shift. From Eq.\ref{CFR2} and Eq.\ref{CFR_Power}, we can see that the total CFR power is the sum of a constant offset and a set of sinusoids, where the frequencies of the sinusoids is a function of the velocity of path length changes. By measuring the frequencies of these sinusoids and multiplying them with the carrier wavelength, we can obtain the speeds of path length change. In this way, we can build a CSI-speed model which relates the variations in CSI power to the movement speeds.
    
    \subsection{Three Dimensional Convolution Network} 
    
    However, human activity is not a simple object movement, and it has complex shapes and different body parts moving at different speeds. Moreover, signals may be reflected through different paths in complex indoor environments. Therefore, we can't simply use CSI-speed model to map human activity detail. Traditional methods employ Time-Frequency analysis tools, such as Short-Time Fourier Transform(STFT) or Discrete Wavelet Transform(DWT), to separate the speeds of path length change in the frequency domain for modelling human activity. Here we will focus on deep learning methods for features extraction, and use 3D-CNN to realize an end-to-end human activity recognition based on CSI temporal signals.
    
    The 3D-CNN is very effective in extracting spatial-temporal features from sequential signals for action recognition. The kernels of 3D-CNN are defined as 5-dimensional tensors: $F\in \mathcal{R}^{N\times C\times T\times H\times W}$, where C is the number of input channels, T, H and W are the temporal length, height and width of these kernels, and N is the number of output channels. The input video volume or internal feature volume is defined as $V \in \mathcal{R}^{C\times L\times X\times Y}$, where L, X and Y are the temporal length and spatial and width of the volume. The operation of each 3D-CNN kernels $F_f\in \mathcal{R}^{C\times T\times H\times W}, f=1,\cdots,N$ is formulated as:
    \begin{equation}
    \begin{split}
    &\Phi_f(l,x,y) = V*F_f \\
    &=\sum_{c=1}^C\sum_{t=1}^T\sum_{h=1}^H\sum_{w=1}^W V(c,l-t,x-h,y-w)F_f(c,t,h,w)
    \end{split}
    \end{equation}
    where $l=1,\cdots,L, x=1,\cdots,X$ and $y=1,\cdots,Y$ are volume of video. 
    
    The C3D net\citep{DBLP:journals/corr/TranBFTP14} is the most widely used 3D-CNN models, which employs traditional $3\times3\times3$ 3D homogeneous convolutional layer. In theory, one can train a C3D model with $3\times3\times3$ kernel as deep as possible to the limitation of the machine memory and computation power. As Fig.\ref{C3D_model}, the C3D model is designed to have 8 convolution layers, 5 pooling layers, followed by two fully connected layers and a softmax output layer. All of 3D convolution kernels are $3\times3\times3$ with stride $1\times1\times$. All 3D pooling layers are $2\times2\times$ except for \emph{pool1} which has kernel size of $1\times2\times2$ and stride $1\times2\times2$ with the intention of preserving the temporal information in the early time. Each fully connected layer has 4096 output units. 
    
    In the meantime, we separately add soft attention mechanism for spatial and temporal features extraction. For the spatial features, we use a full connection layer to train the different weight matrix of each CSI frame pixel. With this weight matrix, the important of CSI frame pixel will be picked up, and assign larger weights to them. For the temporal features, we use the same method to search for important frame. The soft attention mechanism boost the performance of WiFi CSI based human activity recognition.
    
    \begin{figure*}[htbp]
    \centering
    \includegraphics[width=1.0\textwidth]{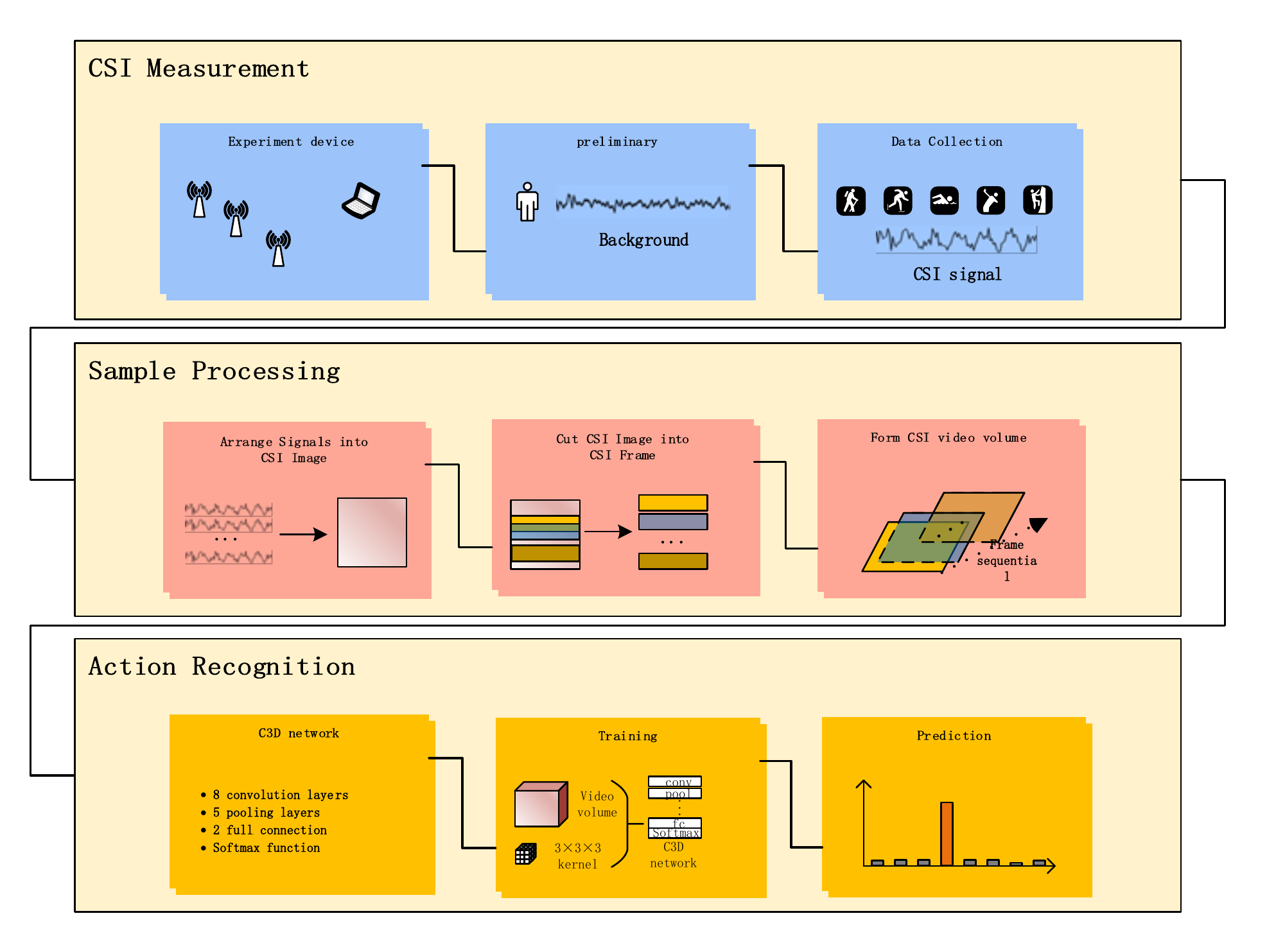}
    \caption{Our model is implemented in three steps: CSI measurement, sample processing and action recognition. In CSI measurement, we collect human action signals with commercial WiFi systems. Then, we process the signal into a picture-like format, and this part will be detailed introduced in following. Last step we will use C3D network to recognize these actions.}
    \label{flowchart}
    \end{figure*} 
    
    Recently, some research propose a kind of effective net. They separate C3D convolutional filter into many cascaded 3D convolutional filters operating on different directions\citep{journals/Pattern/YANG20191}. By this method, the number of net parameters is reduced, and comparing with 2D-CNN case, time comsuming will not be increased so much like C3D model. Here, time is not the main problems for dealing with CSI temporal signals, so we simply use C3D model for our problems. 
    
    \subsection{Our Model}    
    Our approach is mainly based on the two techniques introduced above. As Fig.\ref{flowchart}, the are three steps in our approach. The first step is CSI signals collection(the first row of Fig.\ref{flowchart}), next these CSI signals will be changed into video-volume-like data for C3D net input(the second row of Fig.\ref{flowchart}), and at last C3D net will be used for these CSI signals' recognition(the last row of Fig.\ref{flowchart}).
    
    The signal collection method will be detailed introduce in experiment section of this article. For the second step of our approach, unlike LSTM situation, CSI temporal signals can't be used in C3D model directly, since they are all vector-like features. Therefore, as Fig\ref{flowchart}, we arrange CSI temporal signals into a CSI image. Then, we use a fitting windows to cut CSI frame from this CSI image, and it forms a video-volume-like expression for one human action. As a result, This kind of data can be used in the C3D model.
    
    In the last step, one kind of 3D-CNN network called C3D model will be employed as our recognition model. Since the C3D model is one of the simplest 3D-CNN moel and effective. It also can extract spatial feature and temporal feature at the same time. We do not need another method for CSI signals' denosing and reduce demension. The video-volume-like prepared in the second step will be as input, and fitting $3\times3\times3$ C3D filters will be trained through C3D net and training set. With the help of this model, our proposed approach for human action recognition will be realized. 
       
    \section{experiment}
    
    The experiment is mainly carried out in two parts: first we use commercial Wifi device for CSI signals collection, second we use GPU for C3D network training.
    
    \subsection{CSI Signals Collection Implementation}
    \begin{figure}[htbp]
    \centering
    \includegraphics[width=0.4\textwidth]{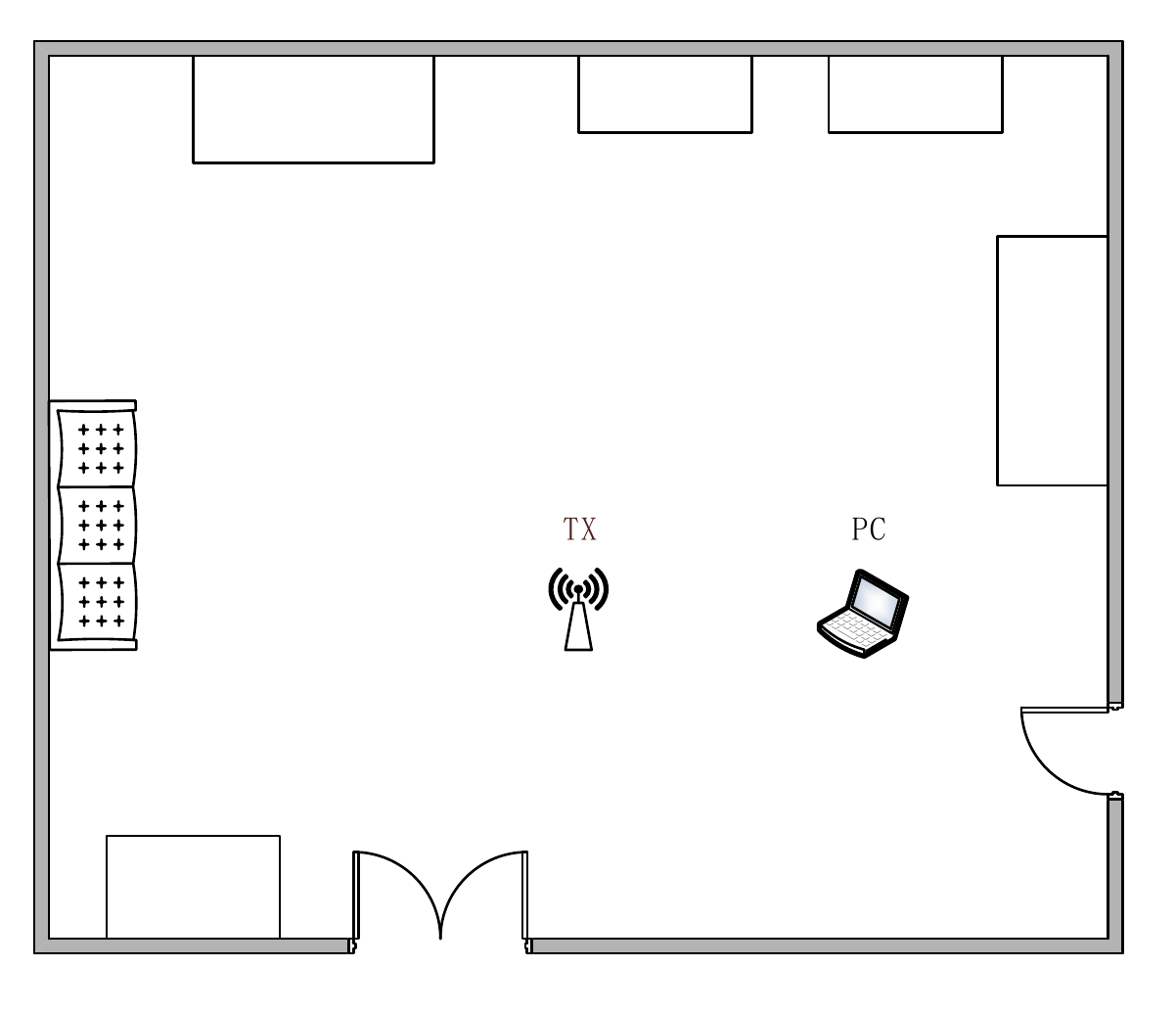}
    \caption{signals collection environment}
    \label{experimental_condition}
    \end{figure} 
    The signals collection device is formed by a commercial wiFi router as a transmitter and a PC with Intel 5300 NIC as a receiver. The sampling frequency is set to 500MHz. There are three transmitter antennas and one receiver antenna in the device. Each antenna owns 30 sub-carriers. Therefore, the raw CSI signals dimension is 90. The transmitter and the receiver are placed three meters apart with line-of-sight(LOS) condition. The collection data comes from 5 persons, and each person performs 14 kinds of activities 20 times. The 14 kinds of activities include bend, draw circle, draw tick, draw x, drink, squat, hand clap, hand up, hand horizontal moving, kick, put off hand, run, sit and walk. During data collection, each person performs each activity for a period of about 20 seconds. Note that, at the beginning and the end of an activity, the person remains stationary. We collected samples for 14 different activities in the lab environment shown in Fig.\ref{experimental_condition}. There are several experiment platform and furniture around the room, and the device is placed in the middle of the room.  
    \begin{figure*}[htbp]
    \centering
    \includegraphics[width=0.9\textwidth]{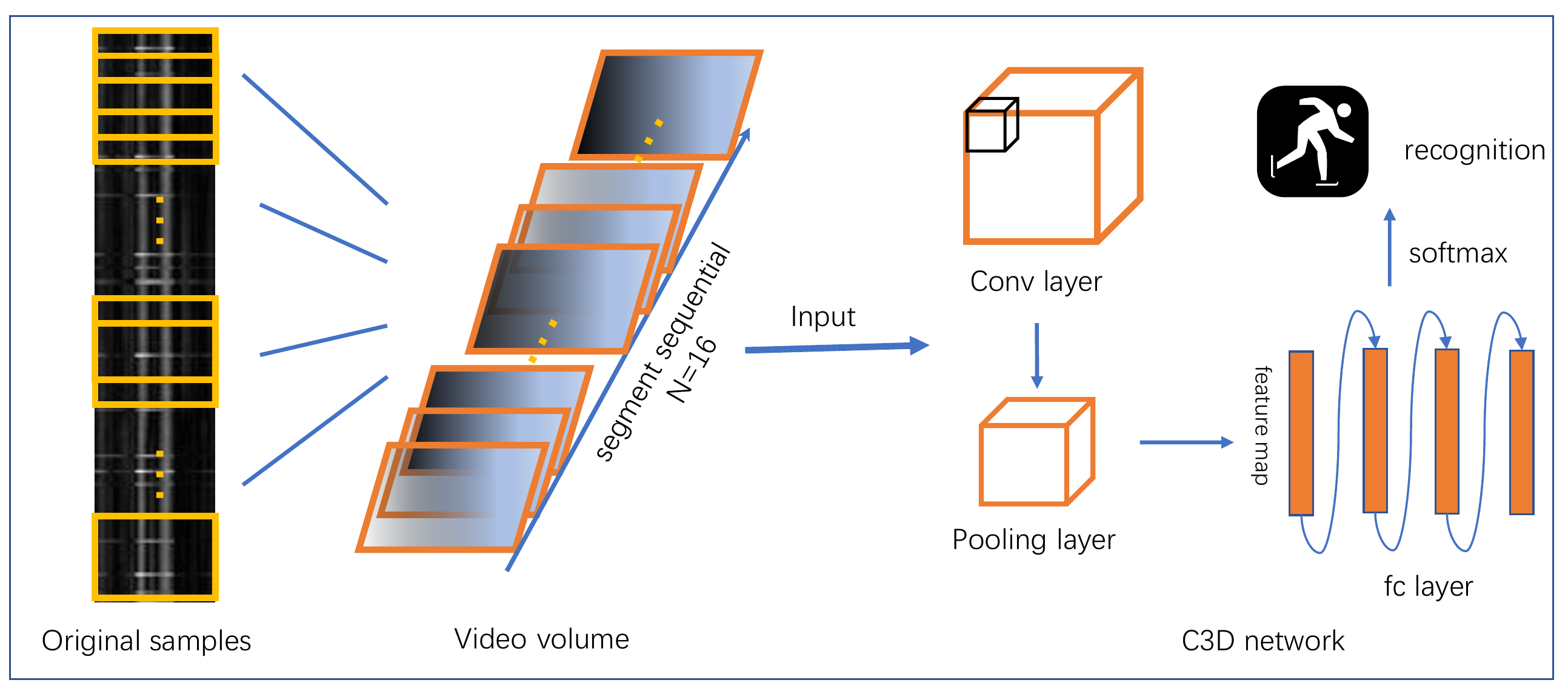}
    \caption{The CSI signals are processed into picture-like data, and by this way they can be put into C3D network.}
    \label{experiment_scheme}
    \end{figure*}  
    
    \subsection{Calculate method}
    
    \begin{figure}[htbp]
    \centering  
    \subfigure[bend]{
    \begin{minipage}[b]{0.23\linewidth}
       \includegraphics[width=1\linewidth]{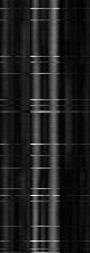}\vspace{4pt}
    \end{minipage}}  
    \subfigure[dun]{
    \begin{minipage}[b]{0.23\linewidth}
       \includegraphics[width=1\linewidth]{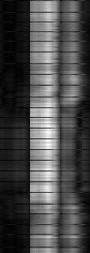}\vspace{4pt}
    \end{minipage}}  
    \subfigure[walk]{
    \begin{minipage}[b]{0.23\linewidth}
       \includegraphics[width=1\linewidth]{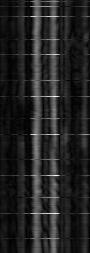}\vspace{4pt}
    \end{minipage}}    
    \caption{signals collection environment}
    \label{CSI_sample}
    \end{figure} 
    
    The experiment calculate part will be finished through C3D network, and the detailed flowchart is shown in Fig.\ref{experiment_scheme}. C3D network can help us for CSI signals' feature extraction and recognition. Before that CSI signals should be designed to meet the C3D network input. Our action's CSI signal data is a $90\times N$ matrix, where N is the activity time of duration, and It's usually around 1000, as shown in Fig.\ref{CSI_sample}. According to the data format, we raise our method in the left part of Fig.\ref{experiment_scheme}: A sliding window(the yellow box in the leftmost part of Fig.\ref{experiment_scheme}) with size of 100 is use for data segmentation, and the stride is set to 8. Therefore, the CSI signals are change into a series of $90\times 100$ CSI images. All of these images are arranged into a video-volume-data. It contains almost $200\sim 300$ frames in the data volume, and therefore it will take a long time to train these original data. Given that, We will random selected clips with 16 frames as our input for C3D network. The size of video-volume-like data for one activity is $16\times 90\times 100$, the separate yellow box in the middle part of the Fig.\ref{experiment_scheme} is just the frame of the video volume. We divide these data equally into training set and test set. The training of the  C3D network is to determine all the model parameters based on the training set with true labels. The test set is used for verifying the training effect of the network. All the experiments are carried out on a testbed equipped with an NVIDIA Geforce GTX 1070Ti GPU card. The duration time for entire training is no more than 10 minutes, and the test time is only about 5 seconds for entire test set, which do not incur much system overhead.
    
   \subsection{result}
   \subsubsection{experiment results}
   To verify the effectiveness of the proposed approach, we perform a comparison with some benchmark approaches for CSI based human activity recognition. Different machine learning techniques have been used for multi-class classification based on certain features that are extracted. Some of the popular classification techniques are Decision Tree(DT), Random Forest(RF), Gradient Boosting Decision Tree(GBDT), support vector machines (SVMs), and k-Nearest Neighbor(KNN) and LSTM. The first 5 techniques are traditional machine learning methods and the last one belongs to deep learning network. However, the data we collected can't be used directly here, and some pretreatment must be done before we put them into these benchmark approaches. For traditional methods, we calculate the means of action temporal sequence according to time variables, and then we use PCA methods to reduce the noise of the feature of these mean values. Through the handling, We can send them into the five traditional machine learning methods. The calculation results are shown in the top half part of Table~\ref{All_result}. We can see that KNN method can give out the highest accuracy. For LSTM method, the sequential samples are just right fit for the LSTM network's input. We only need to use PCA methods to reduce the noises and dimension of the samples, and then send them into LSTM network directly. The results through deep learning methods are shown in the bottom half part of Table~\ref{All_result}. The results shown that the LSTM can get a better result with attention mechanism, and meanwhile they can arrive a higher accuracy comparing with traditional ones.
   \begin{table}[htbp]
   \centering
   \caption{Some methods are compared with C3D result}
   \label{All_result}
   \begin{tabular}{ccc}
   \Xhline{1.2pt}
   Type & Method & Accuracy(\%)\\
   \hline
       & PCA + Decision Tree & 50.28\% \\
       & PCA + Random Forest & 56.14\% \\
   TR  & PCA + GB Decision Tree & 65.86\% \\
       & PCA + SVM & 67.14\% \\
       & PCA + k-NearestNeighbor & 69\% \\
    \hline
       & LSTM & 74.43\% \\
   DL  & Attention + LSTM & 77.57\% \\
       & C3D & \textbf{91.14}\% \\
       & Attention + C3D & \textbf{93.57\%} \\ 
   \Xhline{1.2pt}
   \end{tabular}
   \end{table}
   
   \subsubsection{the C3D methods}
   At the last two lines of Table~\ref{All_result} are the C3D calculation results, whose accuracy is marked with black bold font. We can see that they outperform than other methods, including LSTM. It is an accepted result. For traditional methods, to achieve recognition performance and fit for employed approach, some manual feature extraction for human activity recognition must be done in advance, and just like the temporal sequential average methods used in this article. However, manual feature extraction requires export knowledge. It is also labor intensive and time-consuming. Besides, when the activities to be recognized changed, the designed features may be useless. Moreover, manual features will inevitably miss some implicit key features. In this article, the temporal sequential methods only records the overall trend of movement, and detail  may be omitted. Therefore, some similar actions are hard to recognition.
    
   On the other side, the relationship between temporal CSI signals and human activities is nontrivial. As we all know, LSTM network is very good at dealing with sequential sample. However, our work with CSI signal is quite different from image situation, since the signal at each time node is not equivalent to the each posture during the action in time series CSI signal totally. In table~\ref{All_result}, we can see that the result calculated through LSTM is much higher than traditional method, and it proves that LSTM algorithm outperform in dealing with time series problems. Meanwhile, it's not good enough for the $70\%~80\%$ accuracy. The problem is that the single CSI signal can't stand for one posture perfectly, and directly using the CSI signal as posture's features is not a good choice.
   
   \begin{table}[htbp]
   \centering
   \caption{Confuse matrix}
   \label{confuse_matrix}
   \subtable[LSTM]{
   \resizebox{80mm}{20mm}{
   \begin{tabular}{lcccccccccccccccc}
   \hline
   Activity & \multicolumn{15}{c}{Predicted Labels} \\
   \cline{2-16}
   Labels & BE & DC & DS & DX & DR & SQ & HC & HU & HO & KK & PH & RN & ST & WK & Acc. \\\specialrule{0.05em}{3pt}{3pt}
   BE & \cellcolor[gray]{0.61}39 & 0 & 0 & 0 & \cellcolor[gray]{0.98}2 & \cellcolor[gray]{0.99}1 & \cellcolor[gray]{0.98}2 & 0 & \cellcolor[gray]{0.98}2 & 0 & 0 & 0 & \cellcolor[gray]{0.99}1 & \cellcolor[gray]{0.97}3 & 78 \\
  
   DC & 0 & \cellcolor[gray]{0.67}33 & \cellcolor[gray]{0.95}5 & \cellcolor[gray]{0.98}2 & 0 & 0 & 0 & \cellcolor[gray]{0.97}3 & \cellcolor[gray]{0.95}5 & 0 & \cellcolor[gray]{0.99}1 & 0 & 0 & \cellcolor[gray]{0.99}1 & 66 \\
   
   DS & \cellcolor[gray]{0.99}1 & \cellcolor[gray]{0.94}6 & \cellcolor[gray]{0.68}32 & \cellcolor[gray]{0.95}5 & 0 & \cellcolor[gray]{0.97}3 & 0 & \cellcolor[gray]{0.99}1 & \cellcolor[gray]{0.98}2 & 0 & 0 & 0 & 0 & 0 & 64 \\
   
   DX & 0 & \cellcolor[gray]{0.95}5 & \cellcolor[gray]{0.96}4 & \cellcolor[gray]{0.63}37 & 0 & \cellcolor[gray]{0.97}3 & 0 & 0 & \cellcolor[gray]{0.99}1 & 0 & 0 & 0 & 0 & 0 & 74 \\
   
   DR & 0 & 0 & 0 & \cellcolor[gray]{0.97}3 & \cellcolor[gray]{0.64}36 & \cellcolor[gray]{0.91}9 & 0 & 0 & 0 & \cellcolor[gray]{0.99}1 & 0 & 0 & 0 & \cellcolor[gray]{0.99}1 & 72 \\
   
   SQ & 0 & \cellcolor[gray]{0.99}1 & 0 & 0 & 0 & \cellcolor[gray]{0.55}45 & 0 & 0 & 0 & \cellcolor[gray]{0.97}3 & 0 & 0 & 0 & \cellcolor[gray]{0.99}1 & 90 \\
   
   HC & \cellcolor[gray]{0.99}1 & 0 & 0 & 0 & \cellcolor[gray]{0.99}1 & \cellcolor[gray]{0.94}6 & \cellcolor[gray]{0.61}39 & 0 & 0 & \cellcolor[gray]{0.99}1 & \cellcolor[gray]{0.98}2 & 0 & 0 & 0 & 78 \\
   
   HU & 0 & \cellcolor[gray]{0.95}5 & \cellcolor[gray]{0.98}2 & \cellcolor[gray]{0.98}2 & \cellcolor[gray]{0.99}1 & 0 & \cellcolor[gray]{0.99}1 & \cellcolor[gray]{0.65}35 & \cellcolor[gray]{0.98}2 & \cellcolor[gray]{0.99}1 & \cellcolor[gray]{0.99}1 & 0 & 0 & 0 & 70 \\
   
   HO & 0 & \cellcolor[gray]{0.95}5 & \cellcolor[gray]{0.98}2 & \cellcolor[gray]{0.98}2 & 0 & \cellcolor[gray]{0.99}1 & \cellcolor[gray]{0.98}2 & \cellcolor[gray]{0.96}4 & \cellcolor[gray]{0.72}28 & \cellcolor[gray]{0.96}4 & \cellcolor[gray]{0.99}1 & \cellcolor[gray]{0.99}1 & 0 & 0 & 56 \\
   
   KI & \cellcolor[gray]{0.96}4 & 0 & \cellcolor[gray]{0.96}4 & 0 & \cellcolor[gray]{0.99}1 & \cellcolor[gray]{0.99}1 & \cellcolor[gray]{0.99}1 & \cellcolor[gray]{0.99}1 & 0 & \cellcolor[gray]{0.62}38 & 0 & 0 & 0 & 0 & 76 \\
   
   PH & 0 & \cellcolor[gray]{0.98}2 & 0 & 0 & \cellcolor[gray]{0.99}1 & \cellcolor[gray]{0.99}1 & \cellcolor[gray]{0.95}5 & 0 & \cellcolor[gray]{0.99}1 & \cellcolor[gray]{0.99}1 & \cellcolor[gray]{0.71}39 & 0 & 0 & 0 & 78 \\
   
   RN & 0 & \cellcolor[gray]{0.98}2 & 0 & 0 & 0 & 0 & 0 & 0 & 0 & \cellcolor[gray]{0.99}1 & 0 & \cellcolor[gray]{0.6}40 & 0 & \cellcolor[gray]{0.91}9 & 80 \\
   
   ST & \cellcolor[gray]{0.98}2 & 0 & 0 & 0 & 0 & \cellcolor[gray]{0.99}1 & \cellcolor[gray]{0.99}1 & 0 & 0 & \cellcolor[gray]{0.99}1 & 0 & \cellcolor[gray]{0.99}1 & \cellcolor[gray]{0.59}41 & \cellcolor[gray]{0.97}3 & 82 \\
   
   WK & 0 & 0 & 0 & 0 & 0 & 0 & 0 & 0 & 0 & 0 & 0 & \cellcolor[gray]{0.9}10 & \cellcolor[gray]{0.99}1 & \cellcolor[gray]{0.61}39 & 78 \\
   \hline   
   \end{tabular}}}

   \subtable[LSTM with attention mechanism]{
   \resizebox{80mm}{20mm}{
   \begin{tabular}{lcccccccccccccccc}
   \hline
   Activity & \multicolumn{15}{c}{Predicted Labels} \\
   \cline{2-16}
   Labels & BE & DC & DS & DX & DR & SQ & HC & HU & HO & KK & PH & RN & ST & WK & Acc. \\\specialrule{0.05em}{3pt}{3pt}
   BE & \cellcolor[gray]{0.54}46 & 0 & 0 & 0 & 0 & \cellcolor[gray]{0.99}1 & 0 & 0 & 0 & 0 & 0 & 0 & 0 & \cellcolor[gray]{0.97}3 & 92 \\
   
   DC & 0 & \cellcolor[gray]{0.68}32 & \cellcolor[gray]{0.9}10 & \cellcolor[gray]{0.98}2 & 0 & 0 & 0 & \cellcolor[gray]{0.96}4 & 0 & \cellcolor[gray]{0.99}1 & \cellcolor[gray]{0.99}1 & 0 & 0 & 0 & 64 \\
   
   DS & 0 & 0 & \cellcolor[gray]{0.6}40 & \cellcolor[gray]{0.97}3 & 0 & \cellcolor[gray]{0.98}2 & 0 & \cellcolor[gray]{0.98}2 & \cellcolor[gray]{0.97}3 & 0 & 0 & 0 & 0 & 0 & 80 \\
   
   DX & 0 & \cellcolor[gray]{0.98}2 & \cellcolor[gray]{0.96}4 & \cellcolor[gray]{0.64}36 & 0 & \cellcolor[gray]{0.98}2 & 0 & \cellcolor[gray]{0.98}2 & \cellcolor[gray]{0.97}3 & 0 & 0 & 0 & 0 & \cellcolor[gray]{0.99}1 & 72 \\
  
   DR & 0 & 0 & 0 & \cellcolor[gray]{0.98}2 & \cellcolor[gray]{0.64}36 & \cellcolor[gray]{0.92}8 & 0 & \cellcolor[gray]{0.98}2 & 0 & 0 & \cellcolor[gray]{0.99}1 & 0 & 0 & \cellcolor[gray]{0.99}1 & 72 \\
   
   SQ & 0 & \cellcolor[gray]{0.97}3 & 0 & 0 & 0 & \cellcolor[gray]{0.56}44 & 0 & 0 & 0 & \cellcolor[gray]{0.97}3 & 0 & 0 & 0 & 0 & 88 \\
   
   HC & \cellcolor[gray]{0.98}2 & 0 & 0 & 0 & 0 & \cellcolor[gray]{0.97}3 & \cellcolor[gray]{0.58}42 & 0 & 0 & \cellcolor[gray]{0.98}2 & \cellcolor[gray]{0.88}12 & 0 & 0 & 0 & 84 \\
   
   HU & 0 & \cellcolor[gray]{0.96}4 & 0 & 0 & \cellcolor[gray]{0.99}1 & 0 & 0 & \cellcolor[gray]{0.62}38 & \cellcolor[gray]{0.96}4 & \cellcolor[gray]{0.98}2 & \cellcolor[gray]{0.99}1 & 0 & 0 & 0 & 76 \\
   
   HO & \cellcolor[gray]{0.97}3 & \cellcolor[gray]{0.97}3 & \cellcolor[gray]{0.93}7 & \cellcolor[gray]{0.99}1 & \cellcolor[gray]{0.99}1 & 0 & \cellcolor[gray]{0.98}2 & \cellcolor[gray]{0.9}10 & \cellcolor[gray]{0.77}23 & 0 & 0 & 0 & 0 & 0 & 46 \\
   
   KK & \cellcolor[gray]{0.98}2 & 0 & \cellcolor[gray]{0.98}2 & \cellcolor[gray]{0.99}1 & \cellcolor[gray]{0.99}1 & \cellcolor[gray]{0.98}2 & \cellcolor[gray]{0.98}2 & \cellcolor[gray]{0.99}1 & \cellcolor[gray]{0.99}1 & \cellcolor[gray]{0.63}37 & 0 & 0 & 0 & \cellcolor[gray]{0.99}1 & 74 \\
   
   PH & 0 & 0 & 0 & 0 & \cellcolor[gray]{0.99}1 & \cellcolor[gray]{0.99}1 & \cellcolor[gray]{0.96}4 & \cellcolor[gray]{0.97}3 & 0 & 0 & \cellcolor[gray]{0.59}41 & 0 & 0 & 0 & 82 \\
   
   RN & 0 & 0 & 0 & 0 & 0 & 0 & 0 & 0 & 0 & 0 & 0 & \cellcolor[gray]{0.57}43 & 0 & \cellcolor[gray]{0.93}7 & 86 \\
   
   ST & 0 & 0 & 0 & 0 & 0 & 0 & 0 & 0 & \cellcolor[gray]{0.99}1 & \cellcolor[gray]{0.99}1 & 0 & 0 & \cellcolor[gray]{0.53}47 & \cellcolor[gray]{0.99}1 & 94 \\
   
   WK & 0 & 0 & 0 & 0 & 0 & 0 & 0 & 0 & 0 & 0 & 0 & \cellcolor[gray]{0.88}12 & 0 & \cellcolor[gray]{0.62}38 & 76 \\
   \hline   
   \end{tabular}}}
   
   \subtable[C3D]{
   \resizebox{80mm}{20mm}{
   \begin{tabular}{lcccccccccccccccc}
   \hline
   Activity & \multicolumn{15}{c}{Predicted Labels} \\
   \cline{2-16}
   Labels & BE & DC & DS & DX & DR & SQ & HC & HU & HO & KI & PH & RN & ST & WK & Acc. \\\specialrule{0.05em}{3pt}{3pt}
   BE & \cellcolor[gray]{0.55}45 & 0 & 0 & 0 & \cellcolor[gray]{0.97}3 & 0 & 0 & 0 & 0 & 0 & \cellcolor[gray]{0.98}2 & 0 & 0 & 0 & 90 \\

   DC & 0 & \cellcolor[gray]{0.55}45 & 0 & 0 & 0 & \cellcolor[gray]{0.95}5 & 0 & 0 & 0 & 0 & 0 & 0 & 0 & 0 & 90 \\

   DS & 0 & \cellcolor[gray]{0.95}5 & \cellcolor[gray]{0.55}45 & 0 & 0 & 0 & 0 & 0 & 0 & 0 & 0 & 0 & 0 & 0 & 90 \\

   DX & 0 & 0 & \cellcolor[gray]{0.99}1 & \cellcolor[gray]{0.51}49 & 0 & 0 & 0 & 0 & 0 & 0 & 0 & 0 & 0 & 0 & 98 \\

   DR & 0 & 0 & 0 & \cellcolor[gray]{0.95}5 & \cellcolor[gray]{0.55}45 & 0 & 0 & 0 & 0 & 0 & 0 & 0 & 0 & 0 & 90 \\

   SQ & 0 & 0 & 0 & 0 & 0 & \cellcolor[gray]{0.55}45 & 0 & 0 & 0 & \cellcolor[gray]{0.97}3 & \cellcolor[gray]{0.98}2 & 0 & 0 & 0 & 90 \\

   HC & 0 & 0 & 0 & \cellcolor[gray]{0.95}5 & 0 & 0 & \cellcolor[gray]{0.55}45 & 0 & 0 & 0 & 0 & 0 & 0 & 0 & 90 \\

   HU & 0 & 0 & 0 & \cellcolor[gray]{0.95}5 & 0 & 0 & 0 & \cellcolor[gray]{0.55}45 & 0 & 0 & 0 & 0 & 0 & 0 & 90 \\

   HO & 0 & 0 & 0 & 0 & \cellcolor[gray]{0.95}5 & 0 & 0 & 0 & \cellcolor[gray]{0.55}45 & 0 & 0 & 0 & 0 & 0 & 90 \\

   KK & 0 & 0 & 0 & 0 & 0 & 0 & 0 & 0 & \cellcolor[gray]{0.95}5 & \cellcolor[gray]{0.55}45 & 0 & 0 & 0 & 0 & 90 \\

   PH & 0 & 0 & 0 & 0 & \cellcolor[gray]{0.95}5 & 0 & 0 & 0 & 0 & 0 & \cellcolor[gray]{0.55}45 & 0 & 0 & 0 & 90 \\

   RN & \cellcolor[gray]{0.95}5 & 0 & 0 & 0 & 0 & 0 & 0 & 0 & 0 & 0 & 0 & \cellcolor[gray]{0.55}45 & 0 & 0 & 90 \\

   ST & 0 & 0 & 0 & 0 & 0 & 0 & 0 & 0 & 0 & 0 & 0 & \cellcolor[gray]{0.99}1 & \cellcolor[gray]{0.51}49 & 0 & 98 \\

   WK & \cellcolor[gray]{0.95}5 & 0 & 0 & 0 & 0 & 0 & 0 & 0 & 0 & 0 & 0 & 0 & 0 & \cellcolor[gray]{0.55}45 & 90 \\
   \hline   
   \end{tabular}}}
    
   \subtable[C3D with attention mechanism]{
   \resizebox{80mm}{20mm}{
   \begin{tabular}{lcccccccccccccccc}
   \hline
   Activity & \multicolumn{15}{c}{Predicted Labels} \\
   \cline{2-16}
   Labels & BE & DC & DS & DX & DR & SQ & HC & HU & HO & KK & PH & RN & ST & WK & Acc. \\\specialrule{0.05em}{3pt}{3pt}
   
   BE & \cellcolor[gray]{0.5}50 & 0 & 0 & 0 & 0 & 0 & 0 & 0 & 0 & 0 & 0 & 0 & 0 & 0 & 100 \\

   DC & 0 & \cellcolor[gray]{0.5}50 & 0 & 0 & 0 & 0 & 0 & 0 & 0 & 0 & 0 & 0 & 0 & 0 & 100 \\

   DS & 0 & 0 & \cellcolor[gray]{0.52}48 & 0 & 0 & 0 & \cellcolor[gray]{0.99}1 & \cellcolor[gray]{0.99}1 & 0 & 0 & 0 & 0 & 0 & 0 & 96 \\

   DX & 0 & \cellcolor[gray]{0.99}1 & 0 & \cellcolor[gray]{0.55}45 & 0 & 0 & \cellcolor[gray]{0.99}1 & \cellcolor[gray]{0.99}1 & \cellcolor[gray]{0.99}1 & 0 & 0 & 0 & 0 & 0 & 90 \\

   DR & 0 & 0 & 0 & 0 & \cellcolor[gray]{0.5}50 & 0 & 0 & 0 & 0 & 0 & 0 & 0 & 0 & 0 & 100 \\

   SQ & 0 & 0 & 0 & 0 & 0 & \cellcolor[gray]{0.5}50 & 0 & 0 & 0 & 0 & 0 & 0 & 0 & 0 & 100 \\

   HC & 0 & 0 & \cellcolor[gray]{0.99}1 & 0 & 0 & 0 & \cellcolor[gray]{0.51}49 & 0 & 0 & 0 & 0 & 0 & 0 & 0 & 98 \\

   HU & 0 & 0 & 0 & 0 & 0 & 0 & 0 & \cellcolor[gray]{0.51}49 & \cellcolor[gray]{0.99}1 & 0 & 0 & 0 & 0 & 0 & 98 \\

   HO & 0 & 0 & \cellcolor[gray]{0.92}8 & 0 & 0 & 0 & 0 & \cellcolor[gray]{0.93}7 & \cellcolor[gray]{0.65}35 & 0 & 0 & 0 & 0 & 0 & 70 \\

   KK & 0 & 0 & 0 & 0 & \cellcolor[gray]{0.95}5 & 0 & 0 & 0 & 0 & \cellcolor[gray]{0.55}45 & 0 & 0 & 0 & 0 & 90 \\

   PH & 0 & 0 & 0 & 0 & 0 & 0 & 0 & 0 & 0 & 0 & \cellcolor[gray]{0.5}50 & 0 & 0 & 0 & 100 \\

   RN & \cellcolor[gray]{0.99}1 & 0 & 0 & 0 & 0 & 0 & 0 & 0 & 0 & 0 & 0 & \cellcolor[gray]{0.57}43 & 0 & \cellcolor[gray]{0.94}6 & 86 \\

   ST & 0 & 0 & 0 & 0 & 0 & \cellcolor[gray]{0.95}5 & 0 & 0 & 0 & 0 & 0 & 0 & \cellcolor[gray]{0.56}44 & \cellcolor[gray]{0.99}1 & 88 \\

   WK & 0 & 0 & 0 & 0 & 0 & 0 & 0 & 0 & 0 & 0 & 0 & \cellcolor[gray]{0.98}2 & \cellcolor[gray]{0.97}3 & \cellcolor[gray]{0.55}45 & 90 \\
   \hline   
   \end{tabular}}}
   \end{table}
    
    For this reason, we think C3D may be a better choice than these original CSI signal's feature. The specific process has been discussed in 'calculate method' section. The the CSI signals can be reorganized to get better described action's features through C3D's spatial convolution. The temporal convolution part has a similar function as LSTM, and they can deal with time dimension of CSI signals. As show in Table~\ref{All_result}, the C3D network get the best calculation result. This is a reasonable result, since C3D network contains the traditional method's feature extraction function and LSTM's dealing with time series sample ability. Therefore, C3D network outperforms LSTM in the unintuitive time series problems.
    
    The confusion matrix for deep learning method are show in Table~\ref{confuse_matrix}, and they are LSTM, attention-LSTM, C3D and attention-C3D separately. From confusion matrix, we can see each individual action's accuracy effective on overall accuracy. First of all, for the overall situation, the C3D network outperforms LSTM network. Second, attention mechanism is actually improving the overall accuracy for each network, but some individual action accuracy may get lower. We think it is also affected by the difference between CSI signals features and human truly posture. Some improving on attention mechanism for CSI signals will be done for the future work. Third, some similar action may confuse during the calculation. We design 14 kinds actions, and half of them are related with hands. we can see from the confusion matrix that the hand related action always make confuse with other hands' action. The overall accuracy is good enough, but some detailed feature extraction methods need improving in the future work.
    
    \subsubsection{evaluation}
    We further analyse the recognition result in a statistical view. To comprehensively evaluate the classification result, the following metrics have been widely used: 1) False Positive Rate(FP) indicates the ratio of falsely selected activities as another activity. 2) Precision(PR) is defined as $\frac{TP}{TP+FP}$, where TP is the ratio of a correctly labeled activity. 3) Recall(RE) is $\frac{TP}{TP+FN}$, where FN is the false negative rate. 4) F1-score(F1) is another evaluation metric, defined as $\frac{2*PR*RE}{PR+RE}$. 
    
    \begin{figure}[htbp]
    \centering
    \includegraphics[width=0.5\textwidth]{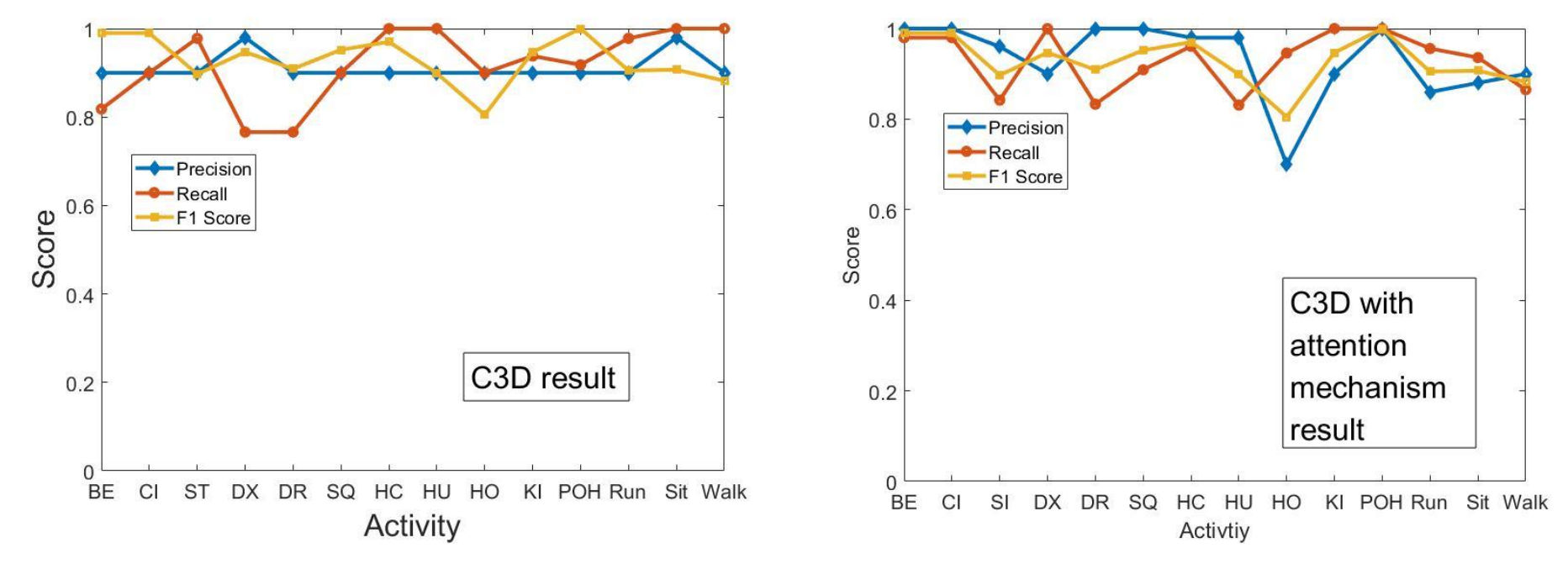}
    \caption{The evaluation result calculated by C3D network and C3D network with attention mechanism.}
    \label{evaluate}
    \end{figure}     
    
    We examine the precision, recall and F1-score for C3D network and C3D network with attention mechanism as illustrated in Fig.~\ref{evaluate}. Since the false positive rate is nearly zero for all actions, we will show them in Fig.~\ref{evaluate}. The precision and recall is almost $90\%$ accuracy except one or two action, and the results for C3D network with attention mechanism get higher accuracy except the hand horizontal moving action. The hand horizontal moving action has a low recognition accuracy, since it is easy to confuse with other hand part action. Meanwhile, it lasts not very long enough time, and has little effect on CSI signal. Therefore, it's also easy to be effected by noise. So some special treating for this action must be done in the future work. 
    
    From evaluate analysis, the result indicates that C3D network and C3D network with attention mechanism can not only accurately but also comprehensively recognize these different activities with low miss and error rates.

    \section{conclusion}
    In this work we try to address the problem of action recognition based on CSI signal features. We proposed to figure out this discrimination model through C3D network. We design some experiments to compare C3D network with traditional machine learning method and LSTM. Traditional machine leaning methods only can extract the CSI signals' features itself, but omit the signals' time direction feature. The other side, LSTM can solve time dimension very well, but is weak to form CSI signals' features. The results showed that C3D network can extract features and make recognition simultaneously. Therefore, it outperforms than all other methods on dealing with these CSI signal data. Totally, the proposed C3D network for dealing with CSI signals are efficient, compact, and extremely simple to use.
    
    \bibliography{refs}

\begin{thebibliography}{44}%
\makeatletter
\providecommand \@ifxundefined [1]{%
 \@ifx{#1\undefined}
}%
\providecommand \@ifnum [1]{%
 \ifnum #1\expandafter \@firstoftwo
 \else \expandafter \@secondoftwo
 \fi
}%
\providecommand \@ifx [1]{%
 \ifx #1\expandafter \@firstoftwo
 \else \expandafter \@secondoftwo
 \fi
}%
\providecommand \natexlab [1]{#1}%
\providecommand \enquote  [1]{``#1''}%
\providecommand \bibnamefont  [1]{#1}%
\providecommand \bibfnamefont [1]{#1}%
\providecommand \citenamefont [1]{#1}%
\providecommand \href@noop [0]{\@secondoftwo}%
\providecommand \href [0]{\begingroup \@sanitize@url \@href}%
\providecommand \@href[1]{\@@startlink{#1}\@@href}%
\providecommand \@@href[1]{\endgroup#1\@@endlink}%
\providecommand \@sanitize@url [0]{\catcode `\\12\catcode `\$12\catcode
  `\&12\catcode `\#12\catcode `\^12\catcode `\_12\catcode `\%12\relax}%
\providecommand \@@startlink[1]{}%
\providecommand \@@endlink[0]{}%
\providecommand \url  [0]{\begingroup\@sanitize@url \@url }%
\providecommand \@url [1]{\endgroup\@href {#1}{\urlprefix }}%
\providecommand \urlprefix  [0]{URL }%
\providecommand \Eprint [0]{\href }%
\providecommand \doibase [0]{http://dx.doi.org/}%
\providecommand \selectlanguage [0]{\@gobble}%
\providecommand \bibinfo  [0]{\@secondoftwo}%
\providecommand \bibfield  [0]{\@secondoftwo}%
\providecommand \translation [1]{[#1]}%
\providecommand \BibitemOpen [0]{}%
\providecommand \bibitemStop [0]{}%
\providecommand \bibitemNoStop [0]{.\EOS\space}%
\providecommand \EOS [0]{\spacefactor3000\relax}%
\providecommand \BibitemShut  [1]{\csname bibitem#1\endcsname}%
\let\auto@bib@innerbib\@empty
\bibitem [{\citenamefont {Aggarwal}\ and\ \citenamefont
  {Ryoo}(2011)}]{journals/csur/AggarwalR11}%
  \BibitemOpen
  \bibfield  {author} {\bibinfo {author} {\bibfnamefont {J.~K.}\ \bibnamefont
  {Aggarwal}}\ and\ \bibinfo {author} {\bibfnamefont {M.~S.}\ \bibnamefont
  {Ryoo}},\ }\href
  {http://dblp.uni-trier.de/db/journals/csur/csur43.html#AggarwalR11}
  {\bibfield  {journal} {\bibinfo  {journal} {ACM Comput. Surv.}\ }\textbf
  {\bibinfo {volume} {43}},\ \bibinfo {pages} {16:1} (\bibinfo {year}
  {2011})}\BibitemShut {NoStop}%
\bibitem [{\citenamefont {{Ji}}\ \emph {et~al.}(2013)\citenamefont {{Ji}},
  \citenamefont {{Xu}}, \citenamefont {{Yang}},\ and\ \citenamefont
  {{Yu}}}]{journals/IEEE/Ji2013}%
  \BibitemOpen
  \bibfield  {author} {\bibinfo {author} {\bibfnamefont {S.}~\bibnamefont
  {{Ji}}}, \bibinfo {author} {\bibfnamefont {W.}~\bibnamefont {{Xu}}}, \bibinfo
  {author} {\bibfnamefont {M.}~\bibnamefont {{Yang}}}, \ and\ \bibinfo {author}
  {\bibfnamefont {K.}~\bibnamefont {{Yu}}},\ }\href {\doibase
  10.1109/TPAMI.2012.59} {\bibfield  {journal} {\bibinfo  {journal} {IEEE
  Transactions on Pattern Analysis and Machine Intelligence}\ }\textbf
  {\bibinfo {volume} {35}},\ \bibinfo {pages} {221} (\bibinfo {year}
  {2013})}\BibitemShut {NoStop}%
\bibitem [{\citenamefont {Schmidhuber}(2015)}]{journals/NN/Schmidhuber201585}%
  \BibitemOpen
  \bibfield  {author} {\bibinfo {author} {\bibfnamefont {J.}~\bibnamefont
  {Schmidhuber}},\ }\href {\doibase
  https://doi.org/10.1016/j.neunet.2014.09.003} {\bibfield  {journal} {\bibinfo
   {journal} {Neural Networks}\ }\textbf {\bibinfo {volume} {61}},\ \bibinfo
  {pages} {85 } (\bibinfo {year} {2015})}\BibitemShut {NoStop}%
\bibitem [{\citenamefont {Vrigkas}\ \emph {et~al.}(2015)\citenamefont
  {Vrigkas}, \citenamefont {Nikou},\ and\ \citenamefont
  {Kakadiaris}}]{journals/FRAI/Vrigkas2015}%
  \BibitemOpen
  \bibfield  {author} {\bibinfo {author} {\bibfnamefont {M.}~\bibnamefont
  {Vrigkas}}, \bibinfo {author} {\bibfnamefont {C.}~\bibnamefont {Nikou}}, \
  and\ \bibinfo {author} {\bibfnamefont {I.}~\bibnamefont {Kakadiaris}},\
  }\href {\doibase 10.3389/frobt.2015.00028} {\bibfield  {journal} {\bibinfo
  {journal} {Frontiers in Robotics and Artificial Intelligence}\ }\textbf
  {\bibinfo {volume} {2}} (\bibinfo {year} {2015}),\
  10.3389/frobt.2015.00028}\BibitemShut {NoStop}%
\bibitem [{\citenamefont {Herath}\ \emph {et~al.}(2017)\citenamefont {Herath},
  \citenamefont {Harandi},\ and\ \citenamefont
  {Porikli}}]{journals/IVC/Herath20174}%
  \BibitemOpen
  \bibfield  {author} {\bibinfo {author} {\bibfnamefont {S.}~\bibnamefont
  {Herath}}, \bibinfo {author} {\bibfnamefont {M.}~\bibnamefont {Harandi}}, \
  and\ \bibinfo {author} {\bibfnamefont {F.}~\bibnamefont {Porikli}},\ }\href
  {\doibase https://doi.org/10.1016/j.imavis.2017.01.010} {\bibfield  {journal}
  {\bibinfo  {journal} {Image and Vision Computing}\ }\textbf {\bibinfo
  {volume} {60}},\ \bibinfo {pages} {4 } (\bibinfo {year} {2017})},\ \bibinfo
  {note} {regularization Techniques for High-Dimensional Data
  Analysis}\BibitemShut {NoStop}%
\bibitem [{\citenamefont {Wang}(2013)}]{journals/PRL/Wang2013}%
  \BibitemOpen
  \bibfield  {author} {\bibinfo {author} {\bibfnamefont {X.}~\bibnamefont
  {Wang}},\ }\href {\doibase https://doi.org/10.1016/j.patrec.2012.07.005}
  {\bibfield  {journal} {\bibinfo  {journal} {Pattern Recognition Letters}\
  }\textbf {\bibinfo {volume} {34}},\ \bibinfo {pages} {3 } (\bibinfo {year}
  {2013})},\ \bibinfo {note} {extracting Semantics from Multi-Spectrum
  Video}\BibitemShut {NoStop}%
\bibitem [{\citenamefont {Rautaray}\ and\ \citenamefont
  {Agrawal}(2015)}]{journals/AIR/Rautaray2015}%
  \BibitemOpen
  \bibfield  {author} {\bibinfo {author} {\bibfnamefont {S.~S.}\ \bibnamefont
  {Rautaray}}\ and\ \bibinfo {author} {\bibfnamefont {A.}~\bibnamefont
  {Agrawal}},\ }\href {\doibase 10.1007/s10462-012-9356-9} {\bibfield
  {journal} {\bibinfo  {journal} {Artificial Intelligence Review}\ }\textbf
  {\bibinfo {volume} {43}},\ \bibinfo {pages} {1} (\bibinfo {year}
  {2015})}\BibitemShut {NoStop}%
\bibitem [{\citenamefont {Wang}\ \emph {et~al.}(2016)\citenamefont {Wang},
  \citenamefont {Wang}, \citenamefont {Zhang}, \citenamefont {Gu},
  \citenamefont {Ni}, \citenamefont {Jia}, \citenamefont {Zhou},\ and\
  \citenamefont {Lv}}]{DBLP:journals/tist/WangWZGNJZL16}%
  \BibitemOpen
  \bibfield  {author} {\bibinfo {author} {\bibfnamefont {T.}~\bibnamefont
  {Wang}}, \bibinfo {author} {\bibfnamefont {Z.}~\bibnamefont {Wang}}, \bibinfo
  {author} {\bibfnamefont {D.}~\bibnamefont {Zhang}}, \bibinfo {author}
  {\bibfnamefont {T.}~\bibnamefont {Gu}}, \bibinfo {author} {\bibfnamefont
  {H.}~\bibnamefont {Ni}}, \bibinfo {author} {\bibfnamefont {J.}~\bibnamefont
  {Jia}}, \bibinfo {author} {\bibfnamefont {X.}~\bibnamefont {Zhou}}, \ and\
  \bibinfo {author} {\bibfnamefont {J.}~\bibnamefont {Lv}},\ }\href {\doibase
  10.1145/2890511} {\bibfield  {journal} {\bibinfo  {journal} {{ACM} {TIST}}\
  }\textbf {\bibinfo {volume} {8}},\ \bibinfo {pages} {6:1} (\bibinfo {year}
  {2016})}\BibitemShut {NoStop}%
\bibitem [{\citenamefont {{Yang}}\ and\ \citenamefont
  {{Tian}}(2017)}]{journals/IEEE/Yang2017}%
  \BibitemOpen
  \bibfield  {author} {\bibinfo {author} {\bibfnamefont {X.}~\bibnamefont
  {{Yang}}}\ and\ \bibinfo {author} {\bibfnamefont {Y.}~\bibnamefont
  {{Tian}}},\ }\href {\doibase 10.1109/TPAMI.2016.2565479} {\bibfield
  {journal} {\bibinfo  {journal} {IEEE Transactions on Pattern Analysis and
  Machine Intelligence}\ }\textbf {\bibinfo {volume} {39}},\ \bibinfo {pages}
  {1028} (\bibinfo {year} {2017})}\BibitemShut {NoStop}%
\bibitem [{\citenamefont {{Mario}}(2019)}]{journals/IEEE/Mario2019}%
  \BibitemOpen
  \bibfield  {author} {\bibinfo {author} {\bibfnamefont {M.}~\bibnamefont
  {{Mario}}},\ }\href {\doibase 10.1109/JSEN.2018.2882943} {\bibfield
  {journal} {\bibinfo  {journal} {IEEE Sensors Journal}\ }\textbf {\bibinfo
  {volume} {19}},\ \bibinfo {pages} {1487} (\bibinfo {year}
  {2019})}\BibitemShut {NoStop}%
\bibitem [{\citenamefont {{Hsieh}}\ and\ \citenamefont
  {{Jeng}}(2018)}]{journals/IEEE/Hsieh2018}%
  \BibitemOpen
  \bibfield  {author} {\bibinfo {author} {\bibfnamefont {Y.}~\bibnamefont
  {{Hsieh}}}\ and\ \bibinfo {author} {\bibfnamefont {Y.}~\bibnamefont
  {{Jeng}}},\ }\href {\doibase 10.1109/ACCESS.2017.2771389} {\bibfield
  {journal} {\bibinfo  {journal} {IEEE Access}\ }\textbf {\bibinfo {volume}
  {6}},\ \bibinfo {pages} {6048} (\bibinfo {year} {2018})}\BibitemShut
  {NoStop}%
\bibitem [{\citenamefont {{Yang}}\ \emph {et~al.}(2019)\citenamefont {{Yang}},
  \citenamefont {{Li}}, \citenamefont {{Yang}},\ and\ \citenamefont
  {{Luo}}}]{journals/IEEE/Yang2019}%
  \BibitemOpen
  \bibfield  {author} {\bibinfo {author} {\bibfnamefont {Z.}~\bibnamefont
  {{Yang}}}, \bibinfo {author} {\bibfnamefont {Y.}~\bibnamefont {{Li}}},
  \bibinfo {author} {\bibfnamefont {J.}~\bibnamefont {{Yang}}}, \ and\ \bibinfo
  {author} {\bibfnamefont {J.}~\bibnamefont {{Luo}}},\ }\href {\doibase
  10.1109/TCSVT.2018.2864148} {\bibfield  {journal} {\bibinfo  {journal} {IEEE
  Transactions on Circuits and Systems for Video Technology}\ }\textbf
  {\bibinfo {volume} {29}},\ \bibinfo {pages} {2405} (\bibinfo {year}
  {2019})}\BibitemShut {NoStop}%
\bibitem [{\citenamefont {Amjadi}\ \emph {et~al.}(2016)\citenamefont {Amjadi},
  \citenamefont {Kyung}, \citenamefont {Park},\ and\ \citenamefont
  {Sitti}}]{journals/AFM/amjadi2015}%
  \BibitemOpen
  \bibfield  {author} {\bibinfo {author} {\bibfnamefont {M.}~\bibnamefont
  {Amjadi}}, \bibinfo {author} {\bibfnamefont {K.-U.}\ \bibnamefont {Kyung}},
  \bibinfo {author} {\bibfnamefont {I.}~\bibnamefont {Park}}, \ and\ \bibinfo
  {author} {\bibfnamefont {M.}~\bibnamefont {Sitti}},\ }\href {\doibase
  10.1002/adfm.201504755} {\bibfield  {journal} {\bibinfo  {journal} {Advanced
  Functional Materials}\ }\textbf {\bibinfo {volume} {26}},\ \bibinfo {pages}
  {1678} (\bibinfo {year} {2016})}\BibitemShut {NoStop}%
\bibitem [{\citenamefont {{Chen}}\ \emph {et~al.}(2017)\citenamefont {{Chen}},
  \citenamefont {{Zhu}}, \citenamefont {{Soh}},\ and\ \citenamefont
  {{Zhang}}}]{journals/IEEE/chen2017}%
  \BibitemOpen
  \bibfield  {author} {\bibinfo {author} {\bibfnamefont {Z.}~\bibnamefont
  {{Chen}}}, \bibinfo {author} {\bibfnamefont {Q.}~\bibnamefont {{Zhu}}},
  \bibinfo {author} {\bibfnamefont {Y.~C.}\ \bibnamefont {{Soh}}}, \ and\
  \bibinfo {author} {\bibfnamefont {L.}~\bibnamefont {{Zhang}}},\ }\href
  {\doibase 10.1109/TII.2017.2712746} {\bibfield  {journal} {\bibinfo
  {journal} {IEEE Transactions on Industrial Informatics}\ }\textbf {\bibinfo
  {volume} {13}},\ \bibinfo {pages} {3070} (\bibinfo {year}
  {2017})}\BibitemShut {NoStop}%
\bibitem [{\citenamefont {{Wang}}\ \emph
  {et~al.}(2017{\natexlab{a}})\citenamefont {{Wang}}, \citenamefont {{Liu}},
  \citenamefont {{Shahzad}}, \citenamefont {{Ling}},\ and\ \citenamefont
  {{Lu}}}]{journals/IEEE/WangLSLL17}%
  \BibitemOpen
  \bibfield  {author} {\bibinfo {author} {\bibfnamefont {W.}~\bibnamefont
  {{Wang}}}, \bibinfo {author} {\bibfnamefont {A.~X.}\ \bibnamefont {{Liu}}},
  \bibinfo {author} {\bibfnamefont {M.}~\bibnamefont {{Shahzad}}}, \bibinfo
  {author} {\bibfnamefont {K.}~\bibnamefont {{Ling}}}, \ and\ \bibinfo {author}
  {\bibfnamefont {S.}~\bibnamefont {{Lu}}},\ }\href {\doibase
  10.1109/JSAC.2017.2679658} {\bibfield  {journal} {\bibinfo  {journal} {IEEE
  Journal on Selected Areas in Communications}\ }\textbf {\bibinfo {volume}
  {35}},\ \bibinfo {pages} {1118} (\bibinfo {year}
  {2017}{\natexlab{a}})}\BibitemShut {NoStop}%
\bibitem [{\citenamefont {{Gu}}\ \emph {et~al.}(2016)\citenamefont {{Gu}},
  \citenamefont {{Ren}},\ and\ \citenamefont {{Li}}}]{journals/IEEE/Gu2016}%
  \BibitemOpen
  \bibfield  {author} {\bibinfo {author} {\bibfnamefont {Y.}~\bibnamefont
  {{Gu}}}, \bibinfo {author} {\bibfnamefont {F.}~\bibnamefont {{Ren}}}, \ and\
  \bibinfo {author} {\bibfnamefont {J.}~\bibnamefont {{Li}}},\ }\href {\doibase
  10.1109/JIOT.2015.2511805} {\bibfield  {journal} {\bibinfo  {journal} {IEEE
  Internet of Things Journal}\ }\textbf {\bibinfo {volume} {3}},\ \bibinfo
  {pages} {796} (\bibinfo {year} {2016})}\BibitemShut {NoStop}%
\bibitem [{\citenamefont {Wang}\ \emph
  {et~al.}(2015{\natexlab{a}})\citenamefont {Wang}, \citenamefont {Liu},
  \citenamefont {Shahzad}, \citenamefont {Ling},\ and\ \citenamefont
  {Lu}}]{proceeding/ACM/wang2015}%
  \BibitemOpen
  \bibfield  {author} {\bibinfo {author} {\bibfnamefont {W.}~\bibnamefont
  {Wang}}, \bibinfo {author} {\bibfnamefont {A.~X.}\ \bibnamefont {Liu}},
  \bibinfo {author} {\bibfnamefont {M.}~\bibnamefont {Shahzad}}, \bibinfo
  {author} {\bibfnamefont {K.}~\bibnamefont {Ling}}, \ and\ \bibinfo {author}
  {\bibfnamefont {S.}~\bibnamefont {Lu}},\ }in\ \href {\doibase
  10.1145/2789168.2790093} {\emph {\bibinfo {booktitle} {Proceedings of the
  21st Annual International Conference on Mobile Computing and Networking}}},\
  \bibinfo {series and number} {MobiCom '15}\ (\bibinfo  {publisher} {ACM},\
  \bibinfo {address} {New York, NY, USA},\ \bibinfo {year} {2015})\ pp.\
  \bibinfo {pages} {65--76}\BibitemShut {NoStop}%
\bibitem [{\citenamefont {Wang}\ \emph {et~al.}(2019)\citenamefont {Wang},
  \citenamefont {Jiang}, \citenamefont {Hou}, \citenamefont {Huang},
  \citenamefont {Dou}, \citenamefont {Zhang},\ and\ \citenamefont
  {Guo}}]{DBLP:journals/access/WangJHHDZG19}%
  \BibitemOpen
  \bibfield  {author} {\bibinfo {author} {\bibfnamefont {Z.}~\bibnamefont
  {Wang}}, \bibinfo {author} {\bibfnamefont {K.}~\bibnamefont {Jiang}},
  \bibinfo {author} {\bibfnamefont {Y.}~\bibnamefont {Hou}}, \bibinfo {author}
  {\bibfnamefont {Z.}~\bibnamefont {Huang}}, \bibinfo {author} {\bibfnamefont
  {W.}~\bibnamefont {Dou}}, \bibinfo {author} {\bibfnamefont {C.}~\bibnamefont
  {Zhang}}, \ and\ \bibinfo {author} {\bibfnamefont {Y.}~\bibnamefont {Guo}},\
  }\href {\doibase 10.1109/ACCESS.2019.2922244} {\bibfield  {journal} {\bibinfo
   {journal} {{IEEE} Access}\ }\textbf {\bibinfo {volume} {7}},\ \bibinfo
  {pages} {78772} (\bibinfo {year} {2019})}\BibitemShut {NoStop}%
\bibitem [{\citenamefont {Al-qaness}\ \emph {et~al.}(2019)\citenamefont
  {Al-qaness}, \citenamefont {Abd~Elaziz}, \citenamefont {Kim}, \citenamefont
  {Ewees}, \citenamefont {Abbasi}, \citenamefont {Alhaj},\ and\ \citenamefont
  {Hawbani}}]{journals/Sensors/Al-qaness2019}%
  \BibitemOpen
  \bibfield  {author} {\bibinfo {author} {\bibfnamefont {M.~A.~A.}\
  \bibnamefont {Al-qaness}}, \bibinfo {author} {\bibfnamefont {M.}~\bibnamefont
  {Abd~Elaziz}}, \bibinfo {author} {\bibfnamefont {S.}~\bibnamefont {Kim}},
  \bibinfo {author} {\bibfnamefont {A.~A.}\ \bibnamefont {Ewees}}, \bibinfo
  {author} {\bibfnamefont {A.~A.}\ \bibnamefont {Abbasi}}, \bibinfo {author}
  {\bibfnamefont {Y.~A.}\ \bibnamefont {Alhaj}}, \ and\ \bibinfo {author}
  {\bibfnamefont {A.}~\bibnamefont {Hawbani}},\ }\href {\doibase
  10.3390/s19153329} {\bibfield  {journal} {\bibinfo  {journal} {Sensors}\
  }\textbf {\bibinfo {volume} {19}} (\bibinfo {year} {2019}),\
  10.3390/s19153329}\BibitemShut {NoStop}%
\bibitem [{\citenamefont {Ma}\ \emph {et~al.}(2019)\citenamefont {Ma},
  \citenamefont {Zhou},\ and\ \citenamefont {Wang}}]{Journals/ACM/Ma2019}%
  \BibitemOpen
  \bibfield  {author} {\bibinfo {author} {\bibfnamefont {Y.}~\bibnamefont
  {Ma}}, \bibinfo {author} {\bibfnamefont {G.}~\bibnamefont {Zhou}}, \ and\
  \bibinfo {author} {\bibfnamefont {S.}~\bibnamefont {Wang}},\ }\href {\doibase
  10.1145/3310194} {\bibfield  {journal} {\bibinfo  {journal} {ACM Comput.
  Surv.}\ }\textbf {\bibinfo {volume} {52}},\ \bibinfo {pages} {46:1} (\bibinfo
  {year} {2019})}\BibitemShut {NoStop}%
\bibitem [{\citenamefont {{Zhang}}\ \emph {et~al.}(2019)\citenamefont
  {{Zhang}}, \citenamefont {{Zhou}}, \citenamefont {{Yang}}, \citenamefont
  {{Ou}},\ and\ \citenamefont {{Xiao}}}]{journals/IEEE/zhang2019}%
  \BibitemOpen
  \bibfield  {author} {\bibinfo {author} {\bibfnamefont {W.}~\bibnamefont
  {{Zhang}}}, \bibinfo {author} {\bibfnamefont {S.}~\bibnamefont {{Zhou}}},
  \bibinfo {author} {\bibfnamefont {L.}~\bibnamefont {{Yang}}}, \bibinfo
  {author} {\bibfnamefont {L.}~\bibnamefont {{Ou}}}, \ and\ \bibinfo {author}
  {\bibfnamefont {Z.}~\bibnamefont {{Xiao}}},\ }\href {\doibase
  10.1109/TVT.2019.2926844} {\bibfield  {journal} {\bibinfo  {journal} {IEEE
  Transactions on Vehicular Technology}\ }\textbf {\bibinfo {volume} {68}},\
  \bibinfo {pages} {7890} (\bibinfo {year} {2019})}\BibitemShut {NoStop}%
\bibitem [{\citenamefont {{Abdelnasser}}\ \emph {et~al.}(2015)\citenamefont
  {{Abdelnasser}}, \citenamefont {{Harras}},\ and\ \citenamefont
  {{Youssef}}}]{inproceedings/IEEE/Abdelnasser2015}%
  \BibitemOpen
  \bibfield  {author} {\bibinfo {author} {\bibfnamefont {H.}~\bibnamefont
  {{Abdelnasser}}}, \bibinfo {author} {\bibfnamefont {K.~A.}\ \bibnamefont
  {{Harras}}}, \ and\ \bibinfo {author} {\bibfnamefont {M.}~\bibnamefont
  {{Youssef}}},\ }in\ \href {\doibase 10.1109/INFCOMW.2015.7179321} {\emph
  {\bibinfo {booktitle} {2015 IEEE Conference on Computer Communications
  Workshops (INFOCOM WKSHPS)}}}\ (\bibinfo {year} {2015})\ pp.\ \bibinfo
  {pages} {17--18}\BibitemShut {NoStop}%
\bibitem [{\citenamefont {{Wang}}\ \emph
  {et~al.}(2017{\natexlab{b}})\citenamefont {{Wang}}, \citenamefont {{Wu}},\
  and\ \citenamefont {{Ni}}}]{journals/IEEE/Wang2017}%
  \BibitemOpen
  \bibfield  {author} {\bibinfo {author} {\bibfnamefont {Y.}~\bibnamefont
  {{Wang}}}, \bibinfo {author} {\bibfnamefont {K.}~\bibnamefont {{Wu}}}, \ and\
  \bibinfo {author} {\bibfnamefont {L.~M.}\ \bibnamefont {{Ni}}},\ }\href
  {\doibase 10.1109/TMC.2016.2557792} {\bibfield  {journal} {\bibinfo
  {journal} {IEEE Transactions on Mobile Computing}\ }\textbf {\bibinfo
  {volume} {16}},\ \bibinfo {pages} {581} (\bibinfo {year}
  {2017}{\natexlab{b}})}\BibitemShut {NoStop}%
\bibitem [{\citenamefont {Wang}\ \emph {et~al.}(2014)\citenamefont {Wang},
  \citenamefont {Liu}, \citenamefont {Chen}, \citenamefont {Gruteser},
  \citenamefont {Yang},\ and\ \citenamefont {Liu}}]{journals/Mobicom/Wang2014}%
  \BibitemOpen
  \bibfield  {author} {\bibinfo {author} {\bibfnamefont {Y.}~\bibnamefont
  {Wang}}, \bibinfo {author} {\bibfnamefont {J.}~\bibnamefont {Liu}}, \bibinfo
  {author} {\bibfnamefont {Y.}~\bibnamefont {Chen}}, \bibinfo {author}
  {\bibfnamefont {M.}~\bibnamefont {Gruteser}}, \bibinfo {author}
  {\bibfnamefont {J.}~\bibnamefont {Yang}}, \ and\ \bibinfo {author}
  {\bibfnamefont {H.}~\bibnamefont {Liu}},\ }\href {\doibase
  10.1145/2639108.2639143} {\bibfield  {journal} {\bibinfo  {journal}
  {Proceedings of the Annual International Conference on Mobile Computing and
  Networking, MOBICOM}\ } (\bibinfo {year} {2014}),\
  10.1145/2639108.2639143}\BibitemShut {NoStop}%
\bibitem [{\citenamefont {Virmani}\ and\ \citenamefont
  {Shahzad}(2017)}]{DBLP:conf/mobisys/VirmaniS17}%
  \BibitemOpen
  \bibfield  {author} {\bibinfo {author} {\bibfnamefont {A.}~\bibnamefont
  {Virmani}}\ and\ \bibinfo {author} {\bibfnamefont {M.}~\bibnamefont
  {Shahzad}},\ }in\ \href {\doibase 10.1145/3081333.3081340} {\emph {\bibinfo
  {booktitle} {Proceedings of the 15th Annual International Conference on
  Mobile Systems, Applications, and Services, MobiSys'17, Niagara Falls, NY,
  USA, June 19-23, 2017}}}\ (\bibinfo {year} {2017})\ pp.\ \bibinfo {pages}
  {252--264}\BibitemShut {NoStop}%
\bibitem [{\citenamefont {Ordóñez}\ and\ \citenamefont
  {Roggen}(2016)}]{journals/Sensors/Ordonez2016}%
  \BibitemOpen
  \bibfield  {author} {\bibinfo {author} {\bibfnamefont {F.~J.}\ \bibnamefont
  {Ordóñez}}\ and\ \bibinfo {author} {\bibfnamefont {D.}~\bibnamefont
  {Roggen}},\ }\href {\doibase 10.3390/s16010115} {\bibfield  {journal}
  {\bibinfo  {journal} {Sensors}\ }\textbf {\bibinfo {volume} {16}} (\bibinfo
  {year} {2016}),\ 10.3390/s16010115}\BibitemShut {NoStop}%
\bibitem [{\citenamefont {{Ullah}}\ \emph {et~al.}(2019)\citenamefont
  {{Ullah}}, \citenamefont {{Muhammad}}, \citenamefont {{Del Ser}},
  \citenamefont {{Baik}},\ and\ \citenamefont {{de
  Albuquerque}}}]{journals/IEEE/Ullah2019}%
  \BibitemOpen
  \bibfield  {author} {\bibinfo {author} {\bibfnamefont {A.}~\bibnamefont
  {{Ullah}}}, \bibinfo {author} {\bibfnamefont {K.}~\bibnamefont {{Muhammad}}},
  \bibinfo {author} {\bibfnamefont {J.}~\bibnamefont {{Del Ser}}}, \bibinfo
  {author} {\bibfnamefont {S.~W.}\ \bibnamefont {{Baik}}}, \ and\ \bibinfo
  {author} {\bibfnamefont {V.~H.~C.}\ \bibnamefont {{de Albuquerque}}},\ }\href
  {\doibase 10.1109/TIE.2018.2881943} {\bibfield  {journal} {\bibinfo
  {journal} {IEEE Transactions on Industrial Electronics}\ }\textbf {\bibinfo
  {volume} {66}},\ \bibinfo {pages} {9692} (\bibinfo {year}
  {2019})}\BibitemShut {NoStop}%
\bibitem [{\citenamefont {{Wang}}\ and\ \citenamefont
  {{Wang}}(2017)}]{journals/IEEE_conf/wang2017}%
  \BibitemOpen
  \bibfield  {author} {\bibinfo {author} {\bibfnamefont {H.}~\bibnamefont
  {{Wang}}}\ and\ \bibinfo {author} {\bibfnamefont {L.}~\bibnamefont
  {{Wang}}},\ }in\ \href {\doibase 10.1109/CVPR.2017.387} {\emph {\bibinfo
  {booktitle} {2017 IEEE Conference on Computer Vision and Pattern Recognition
  (CVPR)}}}\ (\bibinfo {year} {2017})\ pp.\ \bibinfo {pages}
  {3633--3642}\BibitemShut {NoStop}%
\bibitem [{\citenamefont {Yang}\ \emph {et~al.}(2019)\citenamefont {Yang},
  \citenamefont {Yuan}, \citenamefont {Li}, \citenamefont {Du}, \citenamefont
  {Xing}, \citenamefont {Hu},\ and\ \citenamefont
  {Maybank}}]{journals/Pattern/YANG20191}%
  \BibitemOpen
  \bibfield  {author} {\bibinfo {author} {\bibfnamefont {H.}~\bibnamefont
  {Yang}}, \bibinfo {author} {\bibfnamefont {C.}~\bibnamefont {Yuan}}, \bibinfo
  {author} {\bibfnamefont {B.}~\bibnamefont {Li}}, \bibinfo {author}
  {\bibfnamefont {Y.}~\bibnamefont {Du}}, \bibinfo {author} {\bibfnamefont
  {J.}~\bibnamefont {Xing}}, \bibinfo {author} {\bibfnamefont {W.}~\bibnamefont
  {Hu}}, \ and\ \bibinfo {author} {\bibfnamefont {S.~J.}\ \bibnamefont
  {Maybank}},\ }\href {\doibase https://doi.org/10.1016/j.patcog.2018.07.028}
  {\bibfield  {journal} {\bibinfo  {journal} {Pattern Recognition}\ }\textbf
  {\bibinfo {volume} {85}},\ \bibinfo {pages} {1 } (\bibinfo {year}
  {2019})}\BibitemShut {NoStop}%
\bibitem [{\citenamefont {Núñez}\ \emph {et~al.}(2018)\citenamefont
  {Núñez}, \citenamefont {Cabido}, \citenamefont {Pantrigo}, \citenamefont
  {Montemayor},\ and\ \citenamefont {Vélez}}]{journals/Pattern/NUNEZ201880}%
  \BibitemOpen
  \bibfield  {author} {\bibinfo {author} {\bibfnamefont {J.~C.}\ \bibnamefont
  {Núñez}}, \bibinfo {author} {\bibfnamefont {R.}~\bibnamefont {Cabido}},
  \bibinfo {author} {\bibfnamefont {J.~J.}\ \bibnamefont {Pantrigo}}, \bibinfo
  {author} {\bibfnamefont {A.~S.}\ \bibnamefont {Montemayor}}, \ and\ \bibinfo
  {author} {\bibfnamefont {J.~F.}\ \bibnamefont {Vélez}},\ }\href {\doibase
  https://doi.org/10.1016/j.patcog.2017.10.033} {\bibfield  {journal} {\bibinfo
   {journal} {Pattern Recognition}\ }\textbf {\bibinfo {volume} {76}},\
  \bibinfo {pages} {80 } (\bibinfo {year} {2018})}\BibitemShut {NoStop}%
\bibitem [{\citenamefont {{Hu}}\ \emph {et~al.}(2017)\citenamefont {{Hu}},
  \citenamefont {{Huang}}, \citenamefont {{Hu}},\ and\ \citenamefont
  {{Yang}}}]{inproceedings/IEEE/hu2017}%
  \BibitemOpen
  \bibfield  {author} {\bibinfo {author} {\bibfnamefont {Z.}~\bibnamefont
  {{Hu}}}, \bibinfo {author} {\bibfnamefont {G.}~\bibnamefont {{Huang}}},
  \bibinfo {author} {\bibfnamefont {Y.}~\bibnamefont {{Hu}}}, \ and\ \bibinfo
  {author} {\bibfnamefont {Z.}~\bibnamefont {{Yang}}},\ }in\ \href {\doibase
  10.1109/ICIP.2017.8297114} {\emph {\bibinfo {booktitle} {2017 IEEE
  International Conference on Image Processing (ICIP)}}}\ (\bibinfo {year}
  {2017})\ pp.\ \bibinfo {pages} {4402--4406}\BibitemShut {NoStop}%
\bibitem [{\citenamefont {{Wang}}\ \emph {et~al.}(2016)\citenamefont {{Wang}},
  \citenamefont {{Zou}}, \citenamefont {{Zhou}}, \citenamefont {{Wu}},\ and\
  \citenamefont {{Ni}}}]{jounals/IEEE/wang2016}%
  \BibitemOpen
  \bibfield  {author} {\bibinfo {author} {\bibfnamefont {G.}~\bibnamefont
  {{Wang}}}, \bibinfo {author} {\bibfnamefont {Y.}~\bibnamefont {{Zou}}},
  \bibinfo {author} {\bibfnamefont {Z.}~\bibnamefont {{Zhou}}}, \bibinfo
  {author} {\bibfnamefont {K.}~\bibnamefont {{Wu}}}, \ and\ \bibinfo {author}
  {\bibfnamefont {L.~M.}\ \bibnamefont {{Ni}}},\ }\href {\doibase
  10.1109/TMC.2016.2517630} {\bibfield  {journal} {\bibinfo  {journal} {IEEE
  Transactions on Mobile Computing}\ }\textbf {\bibinfo {volume} {15}},\
  \bibinfo {pages} {2907} (\bibinfo {year} {2016})}\BibitemShut {NoStop}%
\bibitem [{\citenamefont {Wang}\ \emph
  {et~al.}(2015{\natexlab{b}})\citenamefont {Wang}, \citenamefont {Liu},
  \citenamefont {Shahzad}, \citenamefont {Ling},\ and\ \citenamefont
  {Lu}}]{inproceedings/Mobicom/wang2015}%
  \BibitemOpen
  \bibfield  {author} {\bibinfo {author} {\bibfnamefont {W.}~\bibnamefont
  {Wang}}, \bibinfo {author} {\bibfnamefont {A.~X.}\ \bibnamefont {Liu}},
  \bibinfo {author} {\bibfnamefont {M.}~\bibnamefont {Shahzad}}, \bibinfo
  {author} {\bibfnamefont {K.}~\bibnamefont {Ling}}, \ and\ \bibinfo {author}
  {\bibfnamefont {S.}~\bibnamefont {Lu}},\ }in\ \href {\doibase
  10.1145/2789168.2790093} {\emph {\bibinfo {booktitle} {Proceedings of the
  21st Annual International Conference on Mobile Computing and Networking}}},\
  \bibinfo {series and number} {MobiCom '15}\ (\bibinfo  {publisher} {ACM},\
  \bibinfo {address} {New York, NY, USA},\ \bibinfo {year} {2015})\ pp.\
  \bibinfo {pages} {65--76}\BibitemShut {NoStop}%
\bibitem [{\citenamefont {{Wu}}\ \emph {et~al.}(2019)\citenamefont {{Wu}},
  \citenamefont {{Chu}}, \citenamefont {{Yang}}, \citenamefont {{Xiang}},
  \citenamefont {{Zheng}},\ and\ \citenamefont
  {{Huang}}}]{journals/IEEE/wu2019}%
  \BibitemOpen
  \bibfield  {author} {\bibinfo {author} {\bibfnamefont {X.}~\bibnamefont
  {{Wu}}}, \bibinfo {author} {\bibfnamefont {Z.}~\bibnamefont {{Chu}}},
  \bibinfo {author} {\bibfnamefont {P.}~\bibnamefont {{Yang}}}, \bibinfo
  {author} {\bibfnamefont {C.}~\bibnamefont {{Xiang}}}, \bibinfo {author}
  {\bibfnamefont {X.}~\bibnamefont {{Zheng}}}, \ and\ \bibinfo {author}
  {\bibfnamefont {W.}~\bibnamefont {{Huang}}},\ }\href {\doibase
  10.1109/TVT.2018.2878754} {\bibfield  {journal} {\bibinfo  {journal} {IEEE
  Transactions on Vehicular Technology}\ }\textbf {\bibinfo {volume} {68}},\
  \bibinfo {pages} {306} (\bibinfo {year} {2019})}\BibitemShut {NoStop}%
\bibitem [{\citenamefont {{Greff}}\ \emph {et~al.}(2017)\citenamefont
  {{Greff}}, \citenamefont {{Srivastava}}, \citenamefont {{Koutník}},
  \citenamefont {{Steunebrink}},\ and\ \citenamefont
  {{Schmidhuber}}}]{journals/IEEE/Greff2017}%
  \BibitemOpen
  \bibfield  {author} {\bibinfo {author} {\bibfnamefont {K.}~\bibnamefont
  {{Greff}}}, \bibinfo {author} {\bibfnamefont {R.~K.}\ \bibnamefont
  {{Srivastava}}}, \bibinfo {author} {\bibfnamefont {J.}~\bibnamefont
  {{Koutník}}}, \bibinfo {author} {\bibfnamefont {B.~R.}\ \bibnamefont
  {{Steunebrink}}}, \ and\ \bibinfo {author} {\bibfnamefont {J.}~\bibnamefont
  {{Schmidhuber}}},\ }\href {\doibase 10.1109/TNNLS.2016.2582924} {\bibfield
  {journal} {\bibinfo  {journal} {IEEE Transactions on Neural Networks and
  Learning Systems}\ }\textbf {\bibinfo {volume} {28}},\ \bibinfo {pages}
  {2222} (\bibinfo {year} {2017})}\BibitemShut {NoStop}%
\bibitem [{\citenamefont {{Wang}}\ \emph
  {et~al.}(2017{\natexlab{c}})\citenamefont {{Wang}}, \citenamefont {{Zhang}},
  \citenamefont {{Gao}}, \citenamefont {{Yue}},\ and\ \citenamefont
  {{Wang}}}]{journals/IEEE_trans/wang2017}%
  \BibitemOpen
  \bibfield  {author} {\bibinfo {author} {\bibfnamefont {J.}~\bibnamefont
  {{Wang}}}, \bibinfo {author} {\bibfnamefont {X.}~\bibnamefont {{Zhang}}},
  \bibinfo {author} {\bibfnamefont {Q.}~\bibnamefont {{Gao}}}, \bibinfo
  {author} {\bibfnamefont {H.}~\bibnamefont {{Yue}}}, \ and\ \bibinfo {author}
  {\bibfnamefont {H.}~\bibnamefont {{Wang}}},\ }\href {\doibase
  10.1109/TVT.2016.2635161} {\bibfield  {journal} {\bibinfo  {journal} {IEEE
  Transactions on Vehicular Technology}\ }\textbf {\bibinfo {volume} {66}},\
  \bibinfo {pages} {6258} (\bibinfo {year} {2017}{\natexlab{c}})}\BibitemShut
  {NoStop}%
\bibitem [{\citenamefont {{Gao}}\ \emph {et~al.}(2017)\citenamefont {{Gao}},
  \citenamefont {{Wang}}, \citenamefont {{Ma}}, \citenamefont {{Feng}},\ and\
  \citenamefont {{Wang}}}]{journals/IEEE/Gao2017}%
  \BibitemOpen
  \bibfield  {author} {\bibinfo {author} {\bibfnamefont {Q.}~\bibnamefont
  {{Gao}}}, \bibinfo {author} {\bibfnamefont {J.}~\bibnamefont {{Wang}}},
  \bibinfo {author} {\bibfnamefont {X.}~\bibnamefont {{Ma}}}, \bibinfo {author}
  {\bibfnamefont {X.}~\bibnamefont {{Feng}}}, \ and\ \bibinfo {author}
  {\bibfnamefont {H.}~\bibnamefont {{Wang}}},\ }\href {\doibase
  10.1109/TVT.2017.2737553} {\bibfield  {journal} {\bibinfo  {journal} {IEEE
  Transactions on Vehicular Technology}\ }\textbf {\bibinfo {volume} {66}},\
  \bibinfo {pages} {10346} (\bibinfo {year} {2017})}\BibitemShut {NoStop}%
\bibitem [{\citenamefont {{Chen}}\ \emph {et~al.}(2019)\citenamefont {{Chen}},
  \citenamefont {{Zhang}}, \citenamefont {{Jiang}}, \citenamefont {{Cao}},\
  and\ \citenamefont {{Cui}}}]{journals/IEEE/chen2019}%
  \BibitemOpen
  \bibfield  {author} {\bibinfo {author} {\bibfnamefont {Z.}~\bibnamefont
  {{Chen}}}, \bibinfo {author} {\bibfnamefont {L.}~\bibnamefont {{Zhang}}},
  \bibinfo {author} {\bibfnamefont {C.}~\bibnamefont {{Jiang}}}, \bibinfo
  {author} {\bibfnamefont {Z.}~\bibnamefont {{Cao}}}, \ and\ \bibinfo {author}
  {\bibfnamefont {W.}~\bibnamefont {{Cui}}},\ }\href {\doibase
  10.1109/TMC.2018.2878233} {\bibfield  {journal} {\bibinfo  {journal} {IEEE
  Transactions on Mobile Computing}\ }\textbf {\bibinfo {volume} {18}},\
  \bibinfo {pages} {2714} (\bibinfo {year} {2019})}\BibitemShut {NoStop}%
\bibitem [{\citenamefont {Prince}(2012)}]{book/princeCVMLI2012}%
  \BibitemOpen
  \bibfield  {author} {\bibinfo {author} {\bibfnamefont {S.}~\bibnamefont
  {Prince}},\ }\href@noop {} {\emph {\bibinfo {title} {{Computer Vision: Models
  Learning and Inference}}}}\ (\bibinfo  {publisher} {{Cambridge University
  Press}},\ \bibinfo {year} {2012})\BibitemShut {NoStop}%
\bibitem [{\citenamefont {{Karpathy}}\ \emph {et~al.}(2014)\citenamefont
  {{Karpathy}}, \citenamefont {{Toderici}}, \citenamefont {{Shetty}},
  \citenamefont {{Leung}}, \citenamefont {{Sukthankar}},\ and\ \citenamefont
  {{Fei-Fei}}}]{inproceedings/IEEE/Karpathy2014}%
  \BibitemOpen
  \bibfield  {author} {\bibinfo {author} {\bibfnamefont {A.}~\bibnamefont
  {{Karpathy}}}, \bibinfo {author} {\bibfnamefont {G.}~\bibnamefont
  {{Toderici}}}, \bibinfo {author} {\bibfnamefont {S.}~\bibnamefont
  {{Shetty}}}, \bibinfo {author} {\bibfnamefont {T.}~\bibnamefont {{Leung}}},
  \bibinfo {author} {\bibfnamefont {R.}~\bibnamefont {{Sukthankar}}}, \ and\
  \bibinfo {author} {\bibfnamefont {L.}~\bibnamefont {{Fei-Fei}}},\ }in\ \href
  {\doibase 10.1109/CVPR.2014.223} {\emph {\bibinfo {booktitle} {2014 IEEE
  Conference on Computer Vision and Pattern Recognition}}}\ (\bibinfo {year}
  {2014})\ pp.\ \bibinfo {pages} {1725--1732}\BibitemShut {NoStop}%
\bibitem [{\citenamefont {{Yousefi}}\ \emph {et~al.}(2017)\citenamefont
  {{Yousefi}}, \citenamefont {{Narui}}, \citenamefont {{Dayal}}, \citenamefont
  {{Ermon}},\ and\ \citenamefont {{Valaee}}}]{journals/IEEE/yousefi2017}%
  \BibitemOpen
  \bibfield  {author} {\bibinfo {author} {\bibfnamefont {S.}~\bibnamefont
  {{Yousefi}}}, \bibinfo {author} {\bibfnamefont {H.}~\bibnamefont {{Narui}}},
  \bibinfo {author} {\bibfnamefont {S.}~\bibnamefont {{Dayal}}}, \bibinfo
  {author} {\bibfnamefont {S.}~\bibnamefont {{Ermon}}}, \ and\ \bibinfo
  {author} {\bibfnamefont {S.}~\bibnamefont {{Valaee}}},\ }\href {\doibase
  10.1109/MCOM.2017.1700082} {\bibfield  {journal} {\bibinfo  {journal} {IEEE
  Communications Magazine}\ }\textbf {\bibinfo {volume} {55}},\ \bibinfo
  {pages} {98} (\bibinfo {year} {2017})}\BibitemShut {NoStop}%
\bibitem [{\citenamefont {Tran}\ \emph {et~al.}(2014)\citenamefont {Tran},
  \citenamefont {Bourdev}, \citenamefont {Fergus}, \citenamefont {Torresani},\
  and\ \citenamefont {Paluri}}]{DBLP:journals/corr/TranBFTP14}%
  \BibitemOpen
  \bibfield  {author} {\bibinfo {author} {\bibfnamefont {D.}~\bibnamefont
  {Tran}}, \bibinfo {author} {\bibfnamefont {L.~D.}\ \bibnamefont {Bourdev}},
  \bibinfo {author} {\bibfnamefont {R.}~\bibnamefont {Fergus}}, \bibinfo
  {author} {\bibfnamefont {L.}~\bibnamefont {Torresani}}, \ and\ \bibinfo
  {author} {\bibfnamefont {M.}~\bibnamefont {Paluri}},\ }\href
  {http://arxiv.org/abs/1412.0767} {\bibfield  {journal} {\bibinfo  {journal}
  {CoRR}\ }\textbf {\bibinfo {volume} {abs/1412.0767}} (\bibinfo {year}
  {2014})},\ \Eprint {http://arxiv.org/abs/1412.0767} {arXiv:1412.0767}
  \BibitemShut {NoStop}%
\bibitem [{\citenamefont {Carreira}\ and\ \citenamefont
  {Zisserman}(2017)}]{DBLP:journals/corr/CarreiraZ17}%
  \BibitemOpen
  \bibfield  {author} {\bibinfo {author} {\bibfnamefont {J.}~\bibnamefont
  {Carreira}}\ and\ \bibinfo {author} {\bibfnamefont {A.}~\bibnamefont
  {Zisserman}},\ }\href {http://arxiv.org/abs/1705.07750} {\bibfield  {journal}
  {\bibinfo  {journal} {CoRR}\ }\textbf {\bibinfo {volume} {abs/1705.07750}}
  (\bibinfo {year} {2017})},\ \Eprint {http://arxiv.org/abs/1705.07750}
  {arXiv:1705.07750} \BibitemShut {NoStop}%
\bibitem [{\citenamefont {Tran}\ \emph {et~al.}(2017)\citenamefont {Tran},
  \citenamefont {Wang}, \citenamefont {Torresani}, \citenamefont {Ray},
  \citenamefont {LeCun},\ and\ \citenamefont
  {Paluri}}]{DBLP:journals/corr/abs-1711-11248}%
  \BibitemOpen
  \bibfield  {author} {\bibinfo {author} {\bibfnamefont {D.}~\bibnamefont
  {Tran}}, \bibinfo {author} {\bibfnamefont {H.}~\bibnamefont {Wang}}, \bibinfo
  {author} {\bibfnamefont {L.}~\bibnamefont {Torresani}}, \bibinfo {author}
  {\bibfnamefont {J.}~\bibnamefont {Ray}}, \bibinfo {author} {\bibfnamefont
  {Y.}~\bibnamefont {LeCun}}, \ and\ \bibinfo {author} {\bibfnamefont
  {M.}~\bibnamefont {Paluri}},\ }\href {http://arxiv.org/abs/1711.11248}
  {\bibfield  {journal} {\bibinfo  {journal} {CoRR}\ }\textbf {\bibinfo
  {volume} {abs/1711.11248}} (\bibinfo {year} {2017})},\ \Eprint
  {http://arxiv.org/abs/1711.11248} {arXiv:1711.11248} \BibitemShut {NoStop}%
\end{thebibliography}%
    
\end{document}